\documentclass[10pt,twocolumn,letterpaper]{article}

\usepackage{iccv}
\usepackage{times}
\usepackage[utf8]{inputenc} 
\usepackage[T1]{fontenc}    
\usepackage{url}            
\usepackage{booktabs}       
\usepackage{amsfonts}       
\usepackage{nicefrac}       
\usepackage{microtype}      
\usepackage{xcolor,colortbl}
\usepackage{graphicx, multirow, graphbox}
\usepackage{subfigure}
\usepackage{tabularx}
\graphicspath{ {figs/} }
\usepackage{comment}
\usepackage{amsmath,amssymb} 
\usepackage{color}
\usepackage{tikz}
\usepackage{needspace}
\usepackage{enumitem,kantlipsum}
\usepackage{wrapfig,lipsum}
\usepackage{pifont}
\usepackage{bm}
\usepackage{float}
\usepackage[accsupp]{axessibility}  
\usepackage{setspace}

\usepackage[pagebackref=true,breaklinks=true,letterpaper=true,colorlinks,bookmarks=false]{hyperref}

\iccvfinalcopy 

\def\iccvPaperID{7959} 

\newcommand{\cmark}{\tiny\ding{51}}%
\newcommand{\xmark}{\tiny\ding{55}}%

\newcommand{\graph}{\mathit{\cal G}}
\newcommand{\pgraph}{\mathit{\hat{\graph}}}

\newcommand{\structn}{\textsc{GraphN}}
\newcommand{\oracle}{\textsc{Oracle-Zs}}

\newcommand{\suppl}{\textit{Appendix}}

\newcommand\blfootnote[1]{%
  \begingroup
  \renewcommand\thefootnote{}\footnote{#1}%
  \addtocounter{footnote}{-1}%
  \endgroup
}

\newcommand{\std}[1]{{\tiny{$\pm$#1}}}
\newcommand\Tstrut{\rule{0pt}{3ex}}
\newcommand\Bstrut{\rule[-1.3ex]{0pt}{0pt}}

\begin{document}
	
	\title{Generative Compositional Augmentations for Scene Graph Prediction}
	
	\author{Boris Knyazev\textsuperscript{*,1,2}\quad Harm de Vries\textsuperscript{3}\quad C\u{a}t\u{a}lina Cangea\textsuperscript{4}\\ 
	Graham W.~Taylor\textsuperscript{1,2}
    Aaron Courville\textsuperscript{5,6}\quad Eugene Belilovsky\textsuperscript{5,7}
    \vspace{1mm}
	\\ \textsuperscript{1}School of Engineering, University of Guelph\quad \textsuperscript{2}Vector Institute for Artificial Intelligence
	\\ \textsuperscript{3}Element AI\quad \textsuperscript{4}University of Cambridge\quad
	\textsuperscript{5}Mila\quad \textsuperscript{6}Université de Montréal\quad \textsuperscript{7}Concordia University
}
	
	\maketitle

	\begin{abstract}
		Inferring objects and their relationships from an image in the form of a scene graph is useful in many applications at the intersection of vision and language. 
		We consider a challenging problem of compositional generalization that emerges in this task due to a long tail data distribution.
		Current scene graph generation models are trained on a tiny fraction of the distribution corresponding to the most frequent compositions, \eg~<cup, on, table>.
		However, test images might contain zero- and few-shot compositions of objects and relationships, \eg~<cup, on, surfboard>. Despite each of the object categories and the predicate (\eg `on') being frequent in the training data, the models often fail to properly understand such unseen or rare compositions. To improve generalization, it is natural to attempt increasing the diversity of the training distribution. However, in the graph domain this is non-trivial. To that end, we propose a method to synthesize rare yet plausible scene graphs by perturbing real ones. We then propose and empirically study a model based on conditional generative adversarial networks (GANs) that allows us to generate visual features of perturbed scene graphs and learn from them in a joint fashion.
		When evaluated on the Visual Genome dataset, our approach yields marginal, but consistent improvements in zero- and few-shot metrics.
		We analyze the limitations of our approach indicating promising directions for future research.
	\end{abstract}
	
\vspace{-5pt}
\section{Introduction\label{sec:intro}}
\blfootnote{\textsuperscript{*}This work was partially done while the author was an intern at Mila. Correspondence to: \texttt{bknyazev@uoguelph.ca}}

Reasoning about the world in terms of objects and relationships between them is an important aspect of human and machine cognition~\cite{greff2020binding}. 
In our environment, we can often observe frequent compositions such as ``person on a surfboard'' or ``person next to a dog''. When we are faced with a rare or previously unseen composition such as ``dog on a surfboard'', to understand the scene we need to understand the concepts of `person', `dog', `surfboard' and `on'. While such unbiased reasoning about concepts is easy for humans, for machines this task has remained extremely challenging~\cite{atzmon2016learning, johnson2017clevr, bahdanau2018systematic, keysers2019measuring, lake2019compositional}. 
Learning-based models tend to capture spurious statistical correlations in the training data~\cite{arjovsky2019invariant,niu2020counterfactual}, \eg~`person' rather than `dog' has always occurred on a surfboard. When the evaluation is explicitly focused on \textit{compositional generalization} -- ability to recognize novel or rare combinations of objects and relationships -- such models then can fail remarkably~\cite{atzmon2016learning, lu2016visual, tang2020unbiased, knyazev2020graph}.

\begin{figure}[t] 
    \centering
	\includegraphics[width=0.4\textwidth]{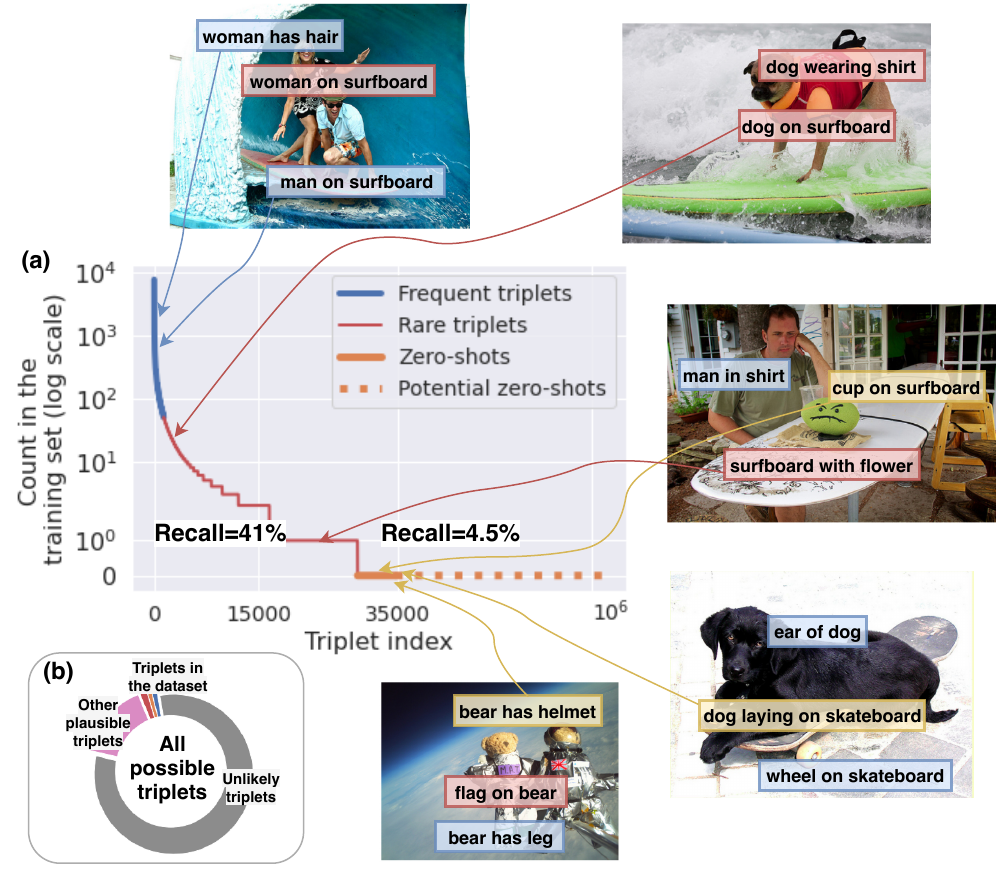}
	\caption{ (\textbf{a}) The triplet distribution in Visual Genome~\cite{krishna2017visual} is extremely long-tailed, with numerous few- and zero-shot compositions (highlighted in red and yellow respectively). (\textbf{b}) The training set contains a tiny fraction (3\%) of all possible triplets, while many other plausible triplets exist. We aim to ``hallucinate'' such compositions using GANs to increase the diversity of training samples and improve generalization. Recall results are from~\cite{tang2020unbiased}.}
	\label{fig:motivation}
	\vspace{-10pt}
\end{figure}

\begin{figure}[t]
\begin{scriptsize}
\setlength{\tabcolsep}{0pt}
\begin{tabular}{cc}
	\includegraphics[width=0.23\textwidth]{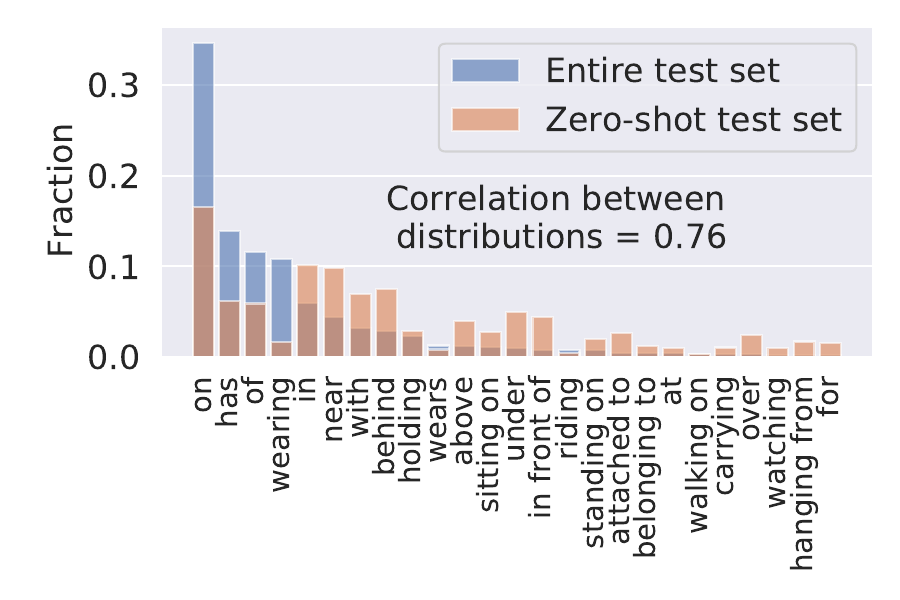} & \includegraphics[width=0.23\textwidth]{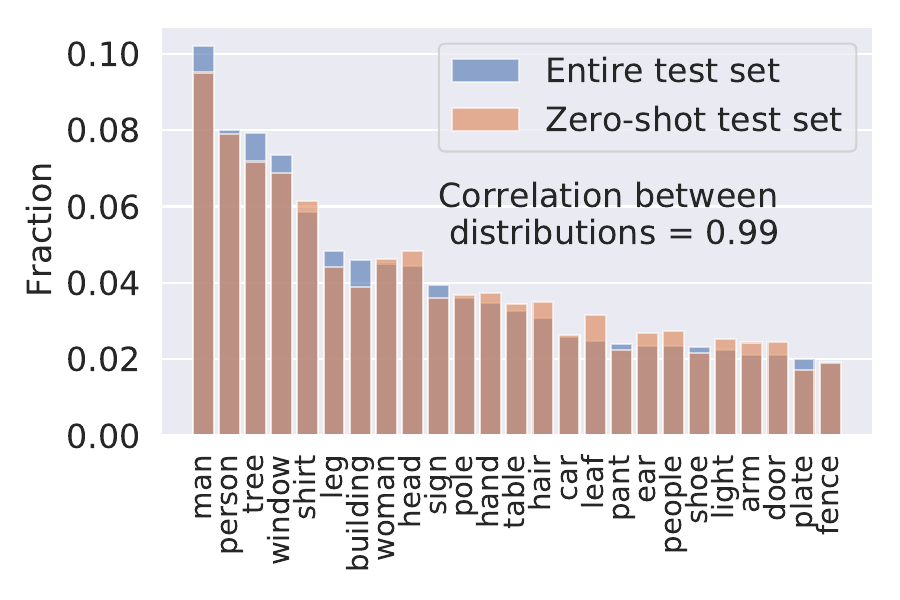} \\
\end{tabular}
\end{scriptsize}
\vspace{-1pt}
\caption{The distributions of top-25 predicate (\textbf{left}) and object (\textbf{right}) categories in Visual Genome~\cite{krishna2017visual} (split of~\cite{xu2017scene}).
}
\vspace{-12pt}
\label{fig:predicates}
\end{figure}

Predicting compositions of objects and the relationships between them from images is part of the scene graph generation (SGG) task. SGG is important, because accurately inferred scene graphs can improve downstream results in tasks, such as VQA~\cite{zhang2019empirical,NSM2019,cangea2019videonavqa,lee2019visual,shi2019explainable,hildebrandt2020scene,damodaran2021understanding}, image captioning~\cite{yang2019auto, gu2019unpaired,li2019know,wang2019role,milewski2020scene}, retrieval~\cite{johnson2015image,belilovsky2017joint,tang2020unbiased,tripathi2019compact,schroeder2020structured} and others~\cite{agarwal2020visual,xu2020survey}.
However, inferring scene graphs accurately is challenging due to a long tail data distribution and inevitable appearance of zero-shot (ZS) compositions (triplets) of objects and relationships at test time, \eg~``cup on surfboard''
(Figure~\ref{fig:motivation}).
The SGG results using the recent Total Direct Effect (TDE) method~\cite{tang2020unbiased} show a severe drop in ZS recall highlighting the extreme challenge of compositional generalization. This might appear surprising given that the marginal distributions in the entire scene graph dataset (\eg~Visual Genome~\cite{krishna2017visual}) and the ZS subset are very similar (Fig.~\ref{fig:predicates}). More specifically, the predicate and object categories that are frequent in the entire dataset, such as `on', `has' and `man', `person' \textit{also dominate} among the ZS triplets. For example, both ``cup on surfboard'' and ``bear has helmet'' consist of frequent entities, but represent extremely rare compositions (Fig.~\ref{fig:motivation}). 
This strongly suggests that the challenging nature of correctly predicting ZS triplets does not directly stem from the imbalance of predicates (or objects), as commonly viewed in the previous SGG works, where the models attempt to improve mean (or predicate-normalized) recall metrics~\cite{chen2019knowledge, dornadula2019visual,tang2019learning,zhang2019graphical,tang2020unbiased,chen2019scene,zareian2020bridging,lin2020gps,zareian2020learning,yan2020pcpl}.
Therefore, we focus on compositional generalization and associated zero- and few-shot metrics.\looseness-1

\begin{figure}
\begin{scriptsize}
\setlength{\tabcolsep}{3.4pt}
\renewcommand{\arraystretch}{1.3}
\begin{tabular}{cccccccc}
	\multicolumn{8}{c}{\includegraphics[width=0.4\textwidth]{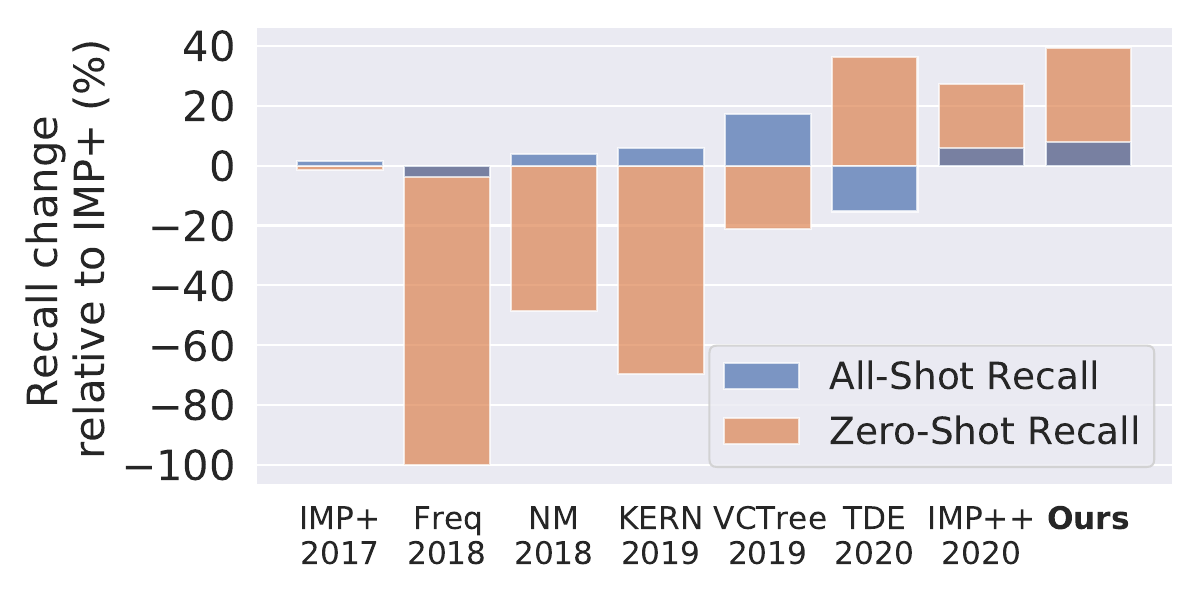}}\vspace{-8pt}\\
	\hspace{48pt} \cite{xu2017scene} & \cite{zellers2018neural} & \cite{zellers2018neural} & \cite{chen2019knowledge} & \cite{tang2020unbiased} & \cite{tang2020unbiased} & \cite{knyazev2020graph} & \hspace{-10pt}{\tiny}\\
\end{tabular}
\end{scriptsize}
\vspace{5pt}
\caption{In this work, the compositional augmentations we propose improve on zero-shot (ZS) as well as all-shot recall.}
\vspace{-5pt}
\label{fig:history}
\end{figure}

Despite recent improvements in compositional generalization within the SGG task~\cite{tang2020unbiased,knyazev2020graph,suhail2021energy}, the state-of-the-art result in zero-shot recall is still 4.5\% compared to 41\% for all-shot recall (Figure~\ref{fig:history}). 
To address compositional generalization, we consider exposing the model to a large diversity of training examples that can lead to emergent generalization~\cite{hill2019environmental,ravuri2019seeing}. To avoid expensive labeling of additional data, we propose a compositional augmentation approach based on conditional generative adversarial networks (GANs)~\cite{goodfellow2014generative,mirza2014conditional}. Our general idea is augmenting the dataset by perturbing scene graphs and corresponding visual features of images, such that together they represent a novel or rare situation. 

Overall, we make the following \textbf{contributions}:
\vspace{-2pt}
\begin{itemize}[labelsep=1pt]
	\vspace{-5pt}
	\itemsep0em
	\item We propose scene graph perturbation methods (\S~\ref{sec:perturb}) as part of a GAN-based model (\S~\ref{sec:model}), to augment the training set with underrepresented compositions;
	\vspace{-3pt}
	\item We propose natural language- and dataset-based metrics to evaluate the quality of (perturbed) scene graphs (\S~\ref{sec:sg_quality});
	\vspace{-3pt}
	\item We extensively evaluate our model and outperform a strong baseline in zero-, few- and all-shot recall (\S~\ref{sec:exper}).
	\vspace{-5pt}
\end{itemize}

Our code is available at {\url{https://github.com/bknyaz/sgg}}.

\begin{figure*}[t]
	\centering
	\vspace{-5pt}
	\centering
	{\includegraphics[width=0.92\textwidth,trim={0 0.3cm 0 0.2cm},clip]{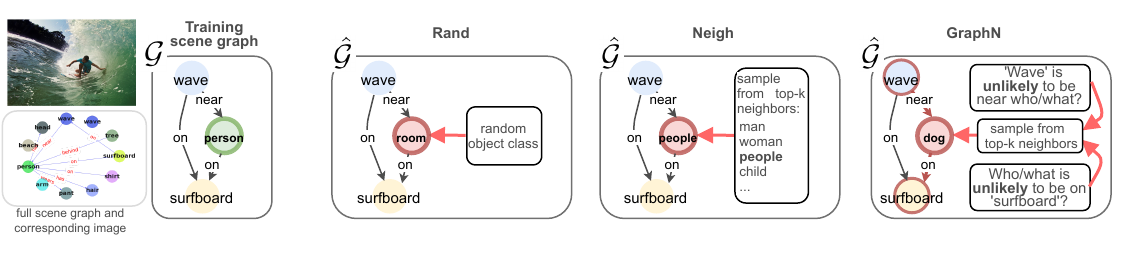}}
	\vspace{-1pt}
	\caption{Illustrative examples of different perturbation schemes we consider. Only the subgraph is shown for clarity. }
	\vspace{-10pt}
	\label{fig:perturb}
\end{figure*}

\section{Related Work}\label{sec:related}
\vspace{-3pt}

\textbf{Scene Graph Generation.} SGG~\cite{xu2017scene} extended an earlier visual relationship detection (VRD) task~\cite{lu2016visual,sadeghi2011recognition}, enabling generation of a complete scene graph (SG) for an image.
This spurred more research at the intersection of vision and language, where a SG can facilitate high-level visual reasoning tasks such as VQA~\cite{zhang2019empirical,NSM2019,shi2019explainable} and others~\cite{agarwal2020visual,xu2020survey,raboh2020differentiable}.
Follow-up SGG works~\cite{li2017scene,yang2018graph, zellers2018neural,zhang2019graphical,gu2019scene,tang2019learning,lu2019learning,lu2021multi} have significantly improved the performance in terms of all-shot recall (Fig.~\ref{fig:history}).
While the problem of zero-shot (ZS) generalization was already actively explored in the VRD task~\cite{zhang2017visual,yang2018shuffle,wang2019generating}, in a more challenging SGG task and on a realistic dataset, such as Visual Genome~\cite{krishna2017visual}, this problem has been addressed only recently in~\cite{tang2020unbiased} 
by proposing Total Direct Effect (TDE), in~\cite{knyazev2020graph} by normalizing the graph loss, and in~\cite{suhail2021energy} by the energy-based loss.
Previous SGG works have not addressed the compositional generalization issue by synthesizing rare SGs.
The closest work that also considers a generative approach is~\cite{wang2019generating} solving the VRD task. Compared to it, our model follows a standard SGG pipeline and evaluation~\cite{xu2017scene,zellers2018neural} including object and predicate classification, instead of classifying only the predicate.
We also condition a GAN on SGs rather than triplets, which combinatorially increases the number of possible augmentations.
To improve SG's likelihood, we leverage both the language model and dataset statistics as opposed to random compositions as in~\cite{wang2019generating}.\looseness-1

\textbf{Predicate imbalance and mean recall.}
Recent SGG works have focused on the predicate imbalance problem~\cite{chen2019knowledge, dornadula2019visual,tang2019learning,zhang2019graphical,tang2020unbiased,chen2019scene,zareian2020bridging,lin2020gps,zareian2020learning,yan2020pcpl} and mean (over predicates) recall as a metric not sensitive to the dominance of frequent predicates. However, as we discussed in \S~\ref{sec:intro}, the challenge of compositional generalization does not directly stem from the imbalance of predicates, since frequent predicates (\eg~`on') still dominate in unseen/rare triplets (Fig.~\ref{fig:predicates}).
Moreover, \cite{tang2020unbiased} showed mean recall is relatively easy to improve by standard Reweight/Resample methods, while ZS recall is not.

\textbf{Data augmentation with GANs.} Data augmentation is a standard method for improving machine learning models \cite{ratner2017learning}. Typically these methods rely on domain specific knowledge such as applying known geometric transformations to images~\cite{devries2017improved,cubuk2018autoaugment}. 
In the case of SGG we require more general augmentation methods, so here we explore a GAN-based approach as one of them.
GANs~\cite{goodfellow2014generative} have been significantly improved w.r.t.~stability of training and the quality of generated samples~\cite{brock2018large,karras2020training}, with recent works considering their usage for data augmentation~\cite{ravuri2019seeing,shin2018medical, sandfort2019data}. Furthermore, recent work has shown that it is possible to produce plausible out-of-distribution (OOD) examples conditioned on unseen label combinations, by intervening on the underlying graph~\cite{kocaoglu2017causalgan,casanova2020generating,sun2020learning,deng2021generative,greff2019multi}. In this work, we have direct access to the underlying graphs of images in the form of SGs, which allows us to condition on OOD compositions as in~\cite{casanova2020generating,deng2021generative}.\looseness-1

\section{Methods}\label{sec:methods}

We consider a dataset of $N$ tuples ${\cal D}=\{(I,\graph,B)\}^N$, where $I$ is an image with a corresponding \textit{scene graph} $\graph$~\cite{johnson2015image} and bounding boxes $B$.
A scene graph $\graph=(O, R)$ consists of $n$ objects $O= \{o_1, ... , o_n\}$, and $m$ relationships between them $R=\{r_1, ..., r_m\}$. 
For each object $o_i$ there is an associated bounding box
$b_i \in \mathbb{R}^{4}, B = \{b_1, ... , b_n\}$.
Each object $o_i$ is labeled with a particular category $o_i \in \cal{C}$, while each relationship $r_k=(i, e_k, j)$ is a triplet with a subject (start node) $i$, an object (end node) $j$ and a predicate $e_k \in {\cal R}$, where $\cal R$ is a set of all predicate classes. 
For further convenience, we define a categorical triplet (\textit{composition})  $\tilde{r}_k=(o_i, e_k, o_j)$ consisting of object and predicate categories, $\tilde{R}=\{\tilde{r}_1, ..., \tilde{r}_m\}$.
An example of a scene graph is presented in Figure~\ref{fig:perturb} with objects $O=\{person, surfboard, wave \}$ and relationships $R=\{ (3,near, 1), (1,on,2) \}$ and categorical relationships $\tilde{R}=\{ (wave,near,person), (person,on,surfboard) \}$.

\subsection{Generative Compositional Augmentations\label{sec:model}}
\vspace{-3pt}

\begin{figure*}[tbph]
	\centering
	\vspace{-5pt}
	{\includegraphics[width=0.8\textwidth, trim={1cm 0cm 2cm 0.5cm}, clip]{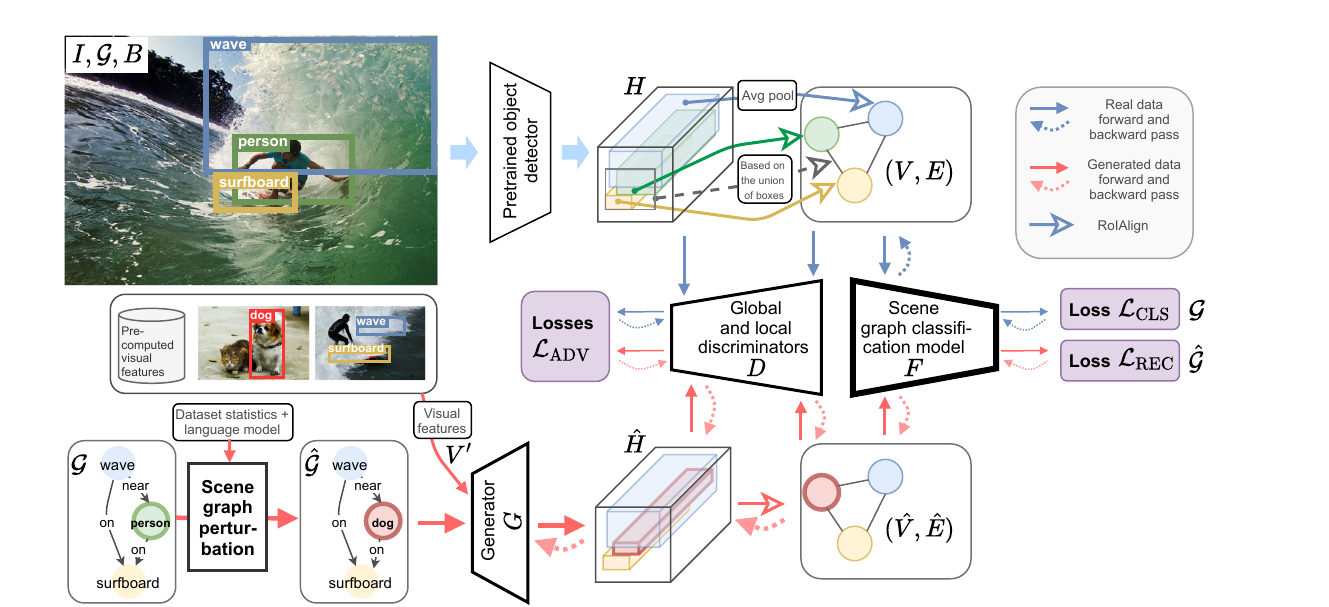}}
	\caption{Our generative scene graph augmentation pipeline with its main components: discriminators $D$, a generator $G$ and a scene graph classification model $F$. See \S~\ref{sec:methods} and \S~\ref{apdx:arch} in \suppl~for a detailed description of our pipeline and model architectures.\looseness-1}
	\vspace{-5pt}
	\label{fig:overview}
\end{figure*}

In a given dataset $\cal D$, such as Visual Genome~\cite{krishna2017visual}, the distribution of triplets is extremely long-tailed with a small fraction of dominating triplets (Fig.~\ref{fig:motivation}). To address the long-tail issue, we consider a GAN-based approach to augment $\cal D$ and artificially upsample rare compositions.
Our model is based on the high-level idea of generating an additional set $\hat{\cal D} = \{ (\hat{I},\pgraph, \hat{B}) \}^{\hat{N}}$. A typical scene-graph-to-image generation pipeline is~\cite{johnson2018image} $\pgraph \rightarrow \hat{B} \rightarrow \hat{I} $. We describe our model accordingly by beginning with constructing $\pgraph$ and $\hat{B}$ (\S~\ref{sec:perturb}) followed by the generation of $\hat{I}$ (in our case, features) (\S~\ref{sec:generation}). See Figure~\ref{fig:overview} for the overall pipeline.

\vspace{-3pt}
\subsubsection{Scene Graph Perturbations\label{sec:perturb}}
\vspace{-3pt}

We propose three methods to synthetically upsample underrepresented triplets in the dataset (Fig.~\ref{fig:perturb}). 
Our goal is to construct diverse compositions avoiding both very likely (already abundant in the dataset) and very unlikely (``implausible'') combinations of objects and predicates, so that the distribution of synthetic $\pgraph$ will resemble the tail of the real distribution of $\graph$.
To construct $\pgraph$, we perturb existing $\graph$ available in $\cal D$, since constructing graphs from scratch is more difficult:
$\graph \rightarrow \pgraph$. We focus on perturbing nodes only as it allows the creation of highly diverse compositions, so $\pgraph=(\hat{O}, R)$, where $\hat{O} = \{ \hat{o}_1, ..., \hat{o}_n\}$ are the replacement object categories. We perturb only $L\cdot n$ nodes, where $L \in \mathbb{R}^{[0,1]}$, so 
$\hat{o}_i = o_i$ for $n (1 - L)$ nodes.
We sample $L\cdot n$ nodes for perturbation based on their sum of in and out degrees. Each scene graph typically has a few ``hub'' nodes densely connected to other nodes.
So, by perturbing the hubs, we introduce more novel compositions with fewer perturbations.\looseness-1

\textbf{\textsc{Rand}} (random) is the simplest strategy, where for a node $i$ we uniformly sample a category $\hat{o}$ from $\cal C$, so that $o_i=\hat{o}$.

\textsc{\textbf{Neigh}} (semantic neighbors) leverages pretrained GloVe word embeddings~\cite{pennington2014glove} available for each of the object categories $\cal C$. Thus, given node $i$ of category $o_i$ we retrieve the top-k neighbors of $o_i$ in the embedding space using cosine similarity. We then uniformly sample $\hat{o}$ from the top-k neighbors replacing $o_i$ with $\hat{o}$.

\textsc{\textbf{\structn}} (graph-structured semantic neighbors). \textsc{Rand} and \textsc{Neigh} do not take into account the graph structure or dataset statistics leading to unlikely or not diverse enough compositions. To alleviate that, we propose the \structn~method. Given node $i$ of category $o_i$ in the graph $\cal G$, we consider all triplets $\tilde{R}_i=\{\tilde{r}_{k,i}\}$ in $\cal G$ that contain $i$ as the start or end node, i.e. $\tilde{r}_{k,i}=(o_i, e_k, o_j) \text{ or } (o_j, e_k, o_i)$. 
For example in Figure~\ref{fig:perturb}, if $o_i$ is `person', then $\tilde{R}_i=\{ (person,on,surfboard), (wave,near,person)\}$.
For each $\tilde{r}_{k,i}$ we find all triplets $\tilde{R}_c$ in the dataset $\cal D$ matching $(o_c, e_k, o_j)$ or $(o_j, e_k, o_c)$, where $o_c \neq o_i$ is a candidate replacement for $o_i$.
For each candidate $o_c$, we count matched triplets $n_c=|\tilde{R}_c|$ and define unnormalized probabilities $\hat{p}_c$ based on the inverse of $n_c$, namely $\hat{p}_c=1/n_c$.
This way we define a set of possible replacements $\{o_c, \hat{p}_c \}$ for node $i$. 

One of our key observations is that depending on the evaluation metric and amount of noise in the dataset, we might want to avoid sampling candidates with very high $\hat{p}_c$ (low $n_c$).
Therefore, to control for that, we introduce an additional hyperparameter $\alpha$ that allows to filter out candidates with $n_c < \alpha$ by setting their $\hat{p}_c$ to 0. 
This way we can trade-off between upsampling rare and frequent triplets.
We then normalize $p_c$ to ensure $\sum p_c = 1$ and sample $o^\prime \sim p_c$. To further increase the diversity, the final $\hat{o}$ is chosen from the top-k semantic neighbors of $o^\prime$ as in \textsc{Neigh}, including $o^\prime$ itself.
\structn~is a sequential perturbation procedure, where for each node the perturbation is conditioned on the current graph state. In contrast, \textsc{Rand} and \textsc{Neigh} perturb all $L\cdot n$ nodes in parallel.\looseness-1

\textbf{Bounding boxes.} Since we perturb only a few nodes, for simplicity we assume that the perturbed graph has the same bounding boxes $B$: $\hat{B}=B$. While one can reasonably argue that object sizes and positions vary a lot depending on the category, i.e. ``elephant'' is much larger than ``dog'', we can often find instances disproving that, \eg~if a toy ``elephant'' or a drawing of an elephant is present. Empirically we found this approach to work well. Please see \S~\ref{sec:pred_box} in \suppl~for the experiments with predicting $\hat{B}$ conditioned on $\pgraph$.

\subsubsection{Scene Graph to Visual Features\label{sec:generation}}

Given perturbed $(\pgraph, \hat{B})$, the next step in our GAN-based pipeline is to generate visual features (Figure~\ref{fig:overview}). 
To train such a model, we first need to extract real features from the dataset ${\cal D}=\{(I,\graph,B)\}^N$.
Following~\cite{xu2017scene,zellers2018neural}, we use a pretrained and frozen object detector~\cite{ren2015faster} to extract global visual features $H$ from input images. 
Then, given $B$ and $H$, we use RoIAlign~\cite{he2017mask} to extract visual features $(V,E)$ of nodes and edges, respectively. To extract edge features between a pair of nodes, the union of their bounding boxes is used~\cite{zellers2018neural}.
Since we do not update the detector, we do not need to generate images as in scene-graph-to-image models~\cite{johnson2018image}, just intermediate features $\hat{H}, \hat{V}, \hat{E}$.

\textbf{Main scene graph classification model $F$.}
Given extracted $(V,E)$, the main model $F$ predicts a scene graph $\graph=(O,R)$, i.e. it needs to correctly assign object labels $O$ to node features $V$ and predicate classes $R$ to edge features $E$. 
Our pipeline is not constrained to the choice of $F$.

\textbf{Generator $G$.}
Our scene-graph-to-features generator $G$ follows the architecture of~\cite{johnson2018image}. First, a scene graph $\pgraph$ is processed by a graph convolutional network (GCN) to exchange information between nodes and edges. We found it beneficial to concatenate output GCN features of all nodes with visual features $V^\prime$, where $V^\prime$ are sampled from the set $\{V_{o_i}\}$ precomputed at the previous stage and ${o_i}$ is the category of node $i$. 
By conditioning the generator on visual features, the main task of $G$ becomes simply to align and smooth the features appropriately, which we believe is easier than generating visual features from the categorical distribution.
In addition, the randomness of this sampling step injects noise improving the diversity of generated features. 
The generated node features and the bounding boxes $\hat{B}$ are used to construct the layout followed by feature refinement~\cite{johnson2018image} to generate $\hat{H}$.
Afterwards, $(\hat{V}, \hat{E})$ are extracted from $\hat{H}$ the same way as $(V,E)$.\looseness=-1

\textbf{Discriminators $D$.}
We have independent discriminators for nodes and edges, $D_{\text{node}}$ and  $D_{\text{edge}}$, that discriminate real features ($V$, $E$) from fake ones ($\hat{V}$, $\hat{E}$) conditioned on their class as per the CGAN~\cite{mirza2014conditional,radford2015unsupervised}. We add a global discriminator $D_{\text{global}}$ acting on feature maps $H$, which encourages global consistency between nodes and edges.
Thus, $D_{\text{node}}$ and $D_{\text{edge}}$ are trained to match marginal distributions, while $D_{\text{global}}$ is trained to match the joint distribution. The right balance between these discriminators should enable the generation of realistic visual features conditioned on OOD scene graphs. Please see \S~\ref{apdx:arch} in \suppl~for the detailed architectures of $D$ and $G$.\looseness-1

\textbf{Losses.}
To train our generative model, we define several losses. These include the baseline SG classification loss \eqref{eq:baseline} and ones specific to our generative pipeline \eqref{eq:rec}-\eqref{eq:adv_full}. The latter are motivated by a CycleGAN~\cite{zhu2017unpaired} and, similarly, consist of the reconstruction and adversarial losses~\eqref{eq:rec}-\eqref{eq:adv_full}. 

We use an improved \textbf{scene graph classification loss} from~\cite{knyazev2020graph}, which is a sum of the node cross-entropy loss ${\cal L}^{O}$ and graph density-normalized edge cross-entropy loss ${\cal L}^{R}$:
\vspace{-5pt}
\setlength{\abovedisplayskip}{2pt}
\setlength{\belowdisplayskip}{2pt}
\begin{align}
\label{eq:baseline}
{\cal L}_\text{CLS}
&= {\cal L}(F(V, E), \graph) =\nonumber \\
&= {\cal L}^{O}(F( V, E), O) + {\cal L}^{R}(F( V, E), R).
\end{align}

${\cal L}^{R}$ is computed based on the ratio of foreground (annotated) to background (not annotated) edges in a batch of scene graphs~\cite{knyazev2020graph}.
To improve $F$ by training it on augmented features $(\hat{V}, \hat{E})$, we define the \textbf{reconstruction (cycle-consistency) loss} analogous to \eqref{eq:baseline}:
\begin{align}
\label{eq:rec}
{\cal L}_\text{REC} =& {\cal L}(F(G(\pgraph, \hat{B}, V^\prime)), {\pgraph}) =\nonumber \\
=& {\cal L}^{O}(F( \hat{V}, \hat{E}), \hat{O}) + {\cal L}^{R}(F( \hat{V}, \hat{E} ), R).
\end{align}
We do not update $G$ on this loss to prevent its potential undesirable collaboration with $F$.
Instead, to train $G$ as well as $D$, we optimize \textbf{conditional adversarial losses}~\cite{mirza2014conditional}.
We first write these separately for $D$ and $G$ in a general form.
So, for some features $\bm{x}$ and their corresponding class $\bm{y}$:
\begin{align}
\label{eq:adv_D}
\mathcal{L}^D_{\text{ADV}}(\bm{x}, \bm{y}) =& \ \mathbb{E}_{\bm{x} \sim p_{\text{data}}(\bm{x})}[\log D(\bm{x}|\bm{y})] +\nonumber \\
& \ \mathbb{E}_{\pgraph \sim p_{\pgraph}(\pgraph)}[\log (1-D(G(\pgraph)|\bm{y})] \\
\mathcal{L}^G_{\text{ADV}}(\bm{y}) =& \ \mathbb{E}_{\pgraph \sim p_{\pgraph}(\pgraph)}[\log D(G(\pgraph)| \bm{y}) ] .\
\end{align}

We compute these losses for object and edge visual features by using the discriminators $D_{\text{node}}$ and  $D_{\text{edge}}$. This loss is also computed for global features $H$ using $D_{\text{global}}$, so that the total discriminator and generator losses are:
\begin{align}
\label{eq:adv_full}
\mathcal{L}^D_{\text{ADV}} &= \mathcal{L}^D_{\text{ADV}}(V,O) + \mathcal{L}^D_{\text{ADV}}(E,R) + \mathcal{L}^D_{\text{ADV}}(H,\emptyset) \nonumber \\
\mathcal{L}^G_{\text{ADV}} &= \mathcal{L}^G_{\text{ADV}}(O) + \mathcal{L}^G_{\text{ADV}}(R) + \mathcal{L}^G_{\text{ADV}}(\emptyset),
\end{align}
\noindent where $\emptyset$ denotes that our global discriminator is unconditional for simplicity.
Thus, the total loss to minimize is:
\vspace{-5pt}
\begin{align}
\label{eq:total_loss}
\mathcal{L} = \underbrace{{\cal L}_{\text{CLS}}  + {\cal L}_{\text{REC}}}_{\text{update } F} - \gamma(\underbrace{\mathcal{L}^D_{\text{ADV}}}_{\text{update } D} + \underbrace{\mathcal{L}^G_{\text{ADV}}}_{\text{update } G}),
\end{align}
\noindent where the loss weight $\gamma=5$ worked well in our experiments.
Compared to a similar work of~\cite{wang2019generating}, in our model all of its components ($F,D,G$) are learned jointly end-to-end.

\subsection{Semantic Plausibility of Scene Graphs\label{sec:sg_quality}}

\textbf{Language model.} To directly evaluate the quality of perturbations, it is desirable to have some quantitative measure other than downstream SGG performance. We found that a cheap (relative to human evaluation) and effective way to achieve this goal is to use a language model. In particular, we use a pretrained BERT~\cite{devlin2018bert} model and estimate the ``semantic plausibility'' of both ground truth and perturbed scene graphs in the following way.
We create a textual query from a scene graph by concatenating all triplets (in a random order). We then mask out one of the perturbed nodes (in case of $\pgraph$) or a random node (in case of $\graph$) in the triplet, so that BERT can return (unnormalized) likelihood scores for the object category of the masked out token. 
We have also considered using this strategy to create SG perturbations as an alternative to \structn. However, we did not find it effective for obtaining rare scene graphs, since BERT is not grounded to visual concepts and not aware of what is considered ``rare'' in a particular SG dataset. For qualitative evaluation and when BERT scores are averaged over many samples, we found them still useful as a rough measure of SG quality. Please see \S~\ref{apdx:bert} in \suppl~for an example of the BERT-based estimation of scene graph quality.\looseness-1

\textbf{Hit rate}. For perturbed SGs, we compute an additional qualitative metric, which we call the `Hit rate'. Assuming we perturbed $M$ triplets in total for all training SGs, this metric computes the percentage of the triplets matching an actual annotation in an evaluation test subset (zero-, few- or all-shot).\looseness-1

\section{Experiments}
\label{sec:exper}
\vspace{-3pt}
\subsection{Dataset, Models and Hyperparameters\label{sec:settins}}
\vspace{-3pt}
We use a publicly available SGG codebase\footnote{\url{https://github.com/rowanz/neural-motifs}} for evaluation and baseline model implementations.
For the model $F$, we use Iterative Message Passing (IMP+)~\cite{xu2017scene, zellers2018neural} and Neural Motifs (NM)~\cite{zellers2018neural}.
IMP+ shows strong compositional generalization capabilities~\cite{knyazev2020graph} and, therefore is more explored in this work.
We use an improved loss for \eqref{eq:baseline} from~\cite{knyazev2020graph}, so we denote our baselines as IMP++ and NM++. 
We use the default hyperparameters and identical setups for the baseline models without a GAN and our models with a GAN. We borrow the detector Faster-RCNN with the VGG16 backbone pretrained on Visual Genome (VG) from~\cite{zellers2018neural} and use it in all our experiments. We evaluate the models on a standard split of VG~\cite{krishna2017visual}, with the 150 most frequent object classes and 50 predicate classes, introduced in~\cite{xu2017scene}. The training set has 57723 and the test set has 26446 images. Similarly to~\cite{knyazev2020graph,wang2019generating,tang2020unbiased,suhail2021energy}, in addition to the all-shot (all test scene graphs) case, we define zero-shot, 10-shot and 100-shot test subsets.
For each such subset we keep only those triplets in a scene graph that occur 0, 1-10 or 11-100 times during training and remove samples without such triplets, which results in 4519, 9602 and 16528 test scene graphs (and images) respectively. 
We use a held-out validation set of 5000 images for tuning the hyperparameters.

\begin{table*}[tbhp]
	\setlength{\tabcolsep}{5pt}
	\scriptsize
	\centering
	\begin{center}
		\caption{Results on Visual Genome~\cite{krishna2017visual} using models based on IMP++~\cite{knyazev2020graph}. The top-1 result in each column is \textbf{bolded} (ignoring \oracle). \oracle~results are an upper bound estimate of ZS recall obtained by directly using ZS test triplets for perturbations. }
		\label{table:main_results}
		\vspace{2pt}
		\begin{tabular}{lccp{0.1cm}ccp{0.1cm}ccp{0.1cm}ccc}
			& \multicolumn{2}{c}{\textsc{\textbf{Zero-shot Recall}}} & &
			\multicolumn{2}{c}{\textsc{\textbf{10-shot Recall}}} & &
			\multicolumn{2}{c}{\textsc{\textbf{100-shot Recall}}} & & 
			\multicolumn{3}{c}{\textsc{\textbf{All-Shot Recall}}} \\
			\textsc{\textbf{Model}} &
			\multicolumn{1}{c}{\scriptsize{SGCls}} & \multicolumn{1}{c}{\scriptsize{PredCls}} & &
			\multicolumn{1}{c}{\scriptsize{SGCls}} & \multicolumn{1}{c}{\scriptsize{PredCls}} & & \multicolumn{1}{c}{\scriptsize{SGCls}} & \multicolumn{1}{c}{\scriptsize{PredCls}} & & \multicolumn{1}{c}{\scriptsize{SGCls}} & \multicolumn{1}{c}{\scriptsize{PredCls}} &  \multicolumn{1}{c}{\scriptsize{SGCls-mR}} \\
			\cline{2-3}\cline{5-6}
			\toprule
			Baseline (IMP++) & 9.27\std{0.10} & 28.14\std{0.05} & & 21.80\std{0.19} & 42.78\std{0.32} & & 40.42\std{0.02} & 67.78\std{0.07} & & 48.70\std{0.08} & 77.48\std{0.09} & 27.78\std{0.10}\Tstrut\Bstrut\\

			GAN+\structn, $\alpha=2$ & \textbf{9.89}\std{0.15} & 28.90\std{0.14} & & 21.96\std{0.30} & \textbf{43.79}\std{0.27} & & 41.22\std{0.33} & 69.17\std{0.24} & & 50.06\std{0.29} & 78.98\std{0.09} & 27.79\std{0.48}\\
			
			GAN+\structn, $\alpha=5$ & 9.62\std{0.29} & \textbf{29.18}\std{0.33} & & \textbf{22.24}\std{0.11} & 43.74\std{0.10} & & 41.39\std{0.26} & 69.11\std{0.05} & & 50.14\std{0.21} & 78.94\std{0.03} & 27.98\std{0.23}\\
			
			GAN+\structn, $\alpha=10$ & 9.84\std{0.17} & 28.90\std{0.46} & & 22.04\std{0.33} & 43.54\std{0.36} & & 41.46\std{0.15} & 69.13\std{0.24} & & 50.10\std{0.23} & 79.00\std{0.09} & 27.68\std{0.37}\\
			
			GAN+\structn, $\alpha=20$ & 9.65\std{0.15} & 28.68\std{0.28} & & 21.97\std{0.30} & 43.64\std{0.20} & & 41.24\std{0.08} & \textbf{69.31}\std{0.17} & & 49.89\std{0.28} & 78.95\std{0.04} & 27.42\std{0.36}\Bstrut\\ 
			
			\hline\hline
			\textbf{Ablated models} \Tstrut\\
			
			GAN (no perturb.) & 9.25\std{0.20} & 28.66\std{0.35} & & 22.15\std{0.21} & 43.66\std{0.29} & & \textbf{41.58}\std{0.20} & 69.16\std{0.16} & & \textbf{50.38}\std{0.28} & \textbf{79.05}\std{0.08} & 28.17\std{0.08}\\
			
			GAN+\textsc{Rand} & 
			9.71\std{0.09} & 28.71\std{0.40} & & 21.89\std{0.21} & 43.33\std{0.18} & & 41.01\std{0.32} & 68.88\std{0.23} & & 49.83\std{0.32} & 78.84\std{0.10} & 27.45\std{0.48}\\
			
			GAN+\textsc{Neigh} & 
			9.65\std{0.04} & 28.68\std{0.40} & & 21.86\std{0.23} & 43.77\std{0.15} & & 41.25\std{0.35} & 69.07\std{0.09} & & 50.00\std{0.36} & 78.94\std{0.10} & 27.41\std{0.51}\Bstrut \\
			
			\hline\hline
			\textbf{Other baselines} \Tstrut\\
			
			\textsc{Reweight} & 9.58\std{0.14} & 28.27\std{0.22} & & 22.19\std{0.09} & 42.98\std{0.17} & & 40.00\std{0.01} & 65.27\std{0.13} & & 48.13\std{0.10} & 74.68\std{0.13} & \textbf{30.95}\std{0.05}\\
			
			\textsc{Resample}-predicates & 9.13\std{0.06} & 27.77\std{0.10} & & 21.35\std{0.05} & 42.14\std{0.16} & & 39.69\std{0.06} & 66.74\std{0.01} & & 48.23\std{0.10} & 76.59\std{0.05} & 28.44\std{0.38} \\
			
    		\textsc{Resample}-triplets & 8.94\std{0.16} & 27.66\std{0.14} & & 21.65\std{0.10} & 42.60\std{0.17} & & 39.39\std{0.08} & 66.44\std{0.06} & & 47.77\std{0.10} & 76.38\std{0.14} & 27.56\std{0.10} \\
    		
    		TDE & 9.21\std{0.21} & 27.91\std{0.09} & & 21.20\std{0.16} & 41.61\std{0.32} & & 39.72\std{0.10} & 65.40\std{0.21} & & 48.35\std{0.08} & 76.22\std{0.17} & 28.25\std{0.21}\Bstrut\\

			\hline\hline
			\multicolumn{2}{l}{\textbf{\textsc{Oracle} perturbations $\pgraph$}} \Tstrut\\
			
			GAN+\oracle~$\pgraph$ & 
			10.11\std{0.34} & 29.27\std{0.10} & & 22.05\std{0.38} & 43.78\std{0.09} & & 41.38\std{0.50} & 69.06\std{0.16} & & 50.19\std{0.36} & 79.00\std{0.08} & 27.91\std{0.56}\\
			
			GAN+\oracle~$\pgraph + \hat{B}$ & 10.52\std{0.31} & 29.43\std{0.42} & & 21.98\std{0.39} & 43.03\std{0.13} & & 41.12\std{0.19} & 68.73\std{0.17} & & 50.05\std{0.35} & 78.65\std{0.09} & 27.52\std{0.46}\\
			
			\bottomrule
		\end{tabular}
	\end{center}
	\vspace{-10pt}
\end{table*}

\begin{figure*}[htpb]
	\vspace{-5pt}
	\centering
	\small
	\setlength{\tabcolsep}{3.5pt}
	\begin{tabular}{ccccc}
 		& \textbf{(a)} Zero-shot hit rate &  \textbf{(b)} 10-shot hit rate & 
 		\textbf{(c)} 100-shot hit rate & 
 		\textbf{(d)} All-shot hit rate \vspace{-1pt} \\ 
 		{\includegraphics[align=c,width=0.13\textwidth,trim={8cm 3.5cm 1cm 3.5cm},clip]{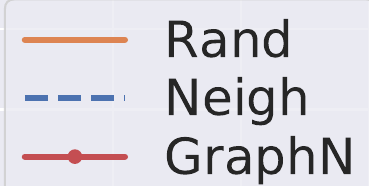}} 
 		&
        \includegraphics[align=c,width=0.2\textwidth,trim={0 0 0 0.5cm},clip]{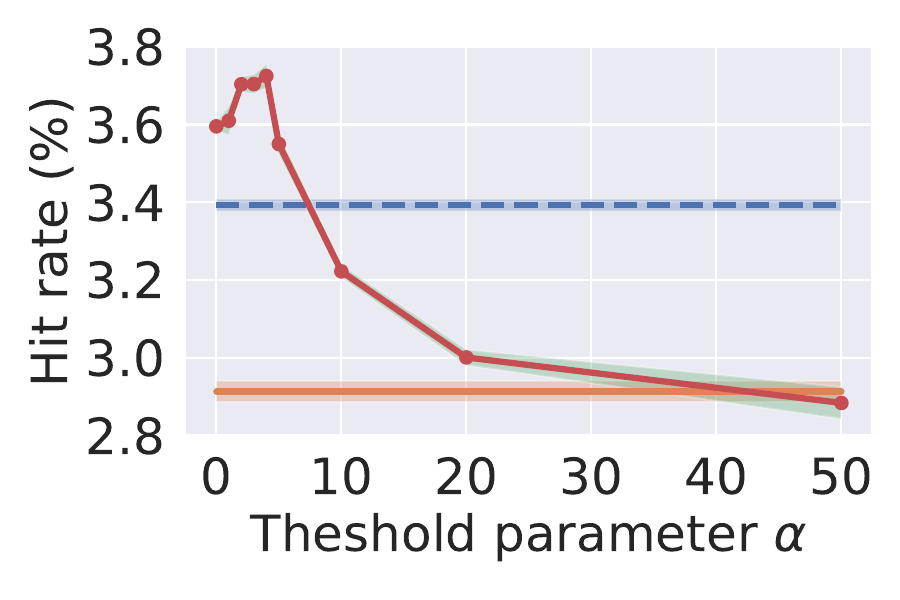} &
        \includegraphics[align=c,width=0.2\textwidth,trim={0 0 0 0.5cm},clip]{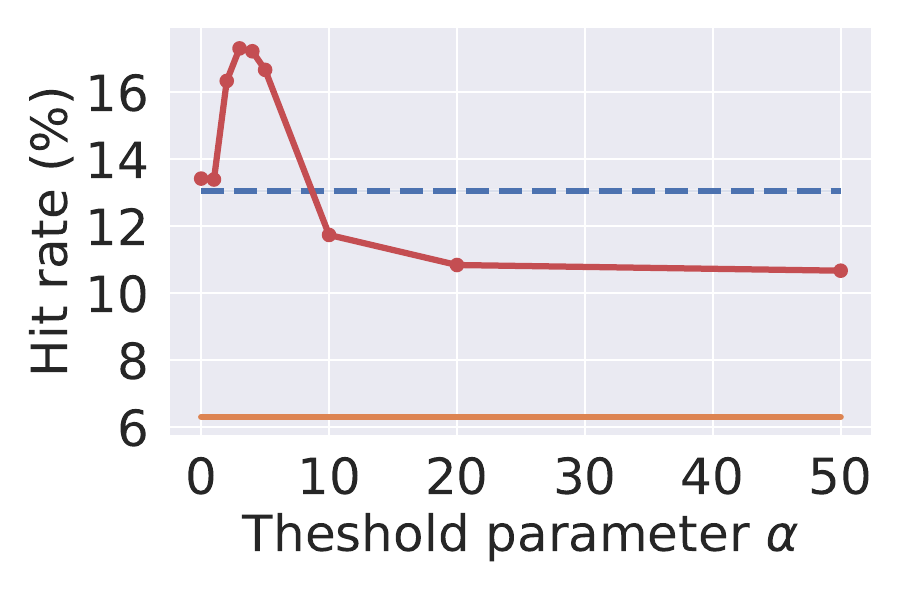} 
        &
        \includegraphics[align=c,width=0.2\textwidth,trim={0 0 0 0.5cm},clip]{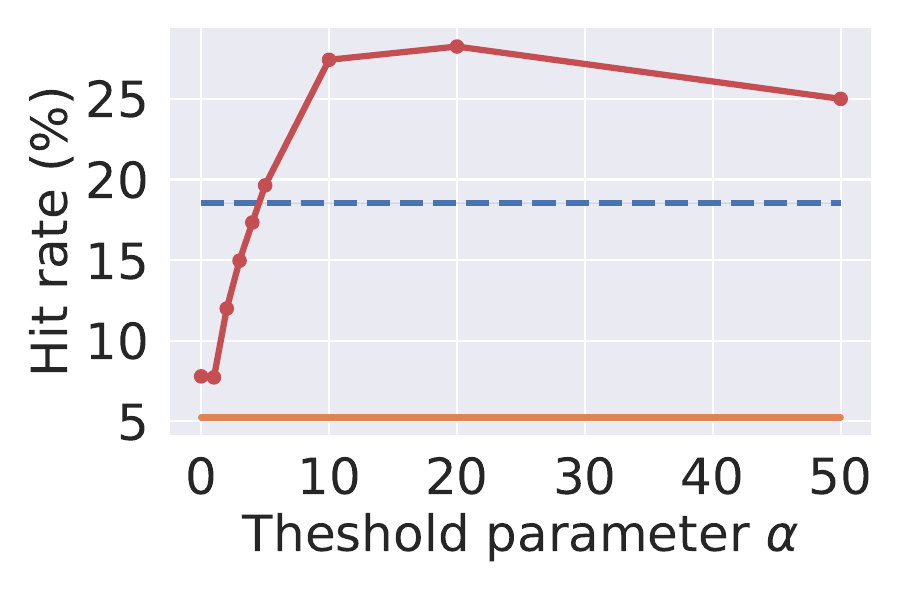} 
        &
        \includegraphics[align=c,width=0.2\textwidth,trim={0 0 0 0.5cm},clip]{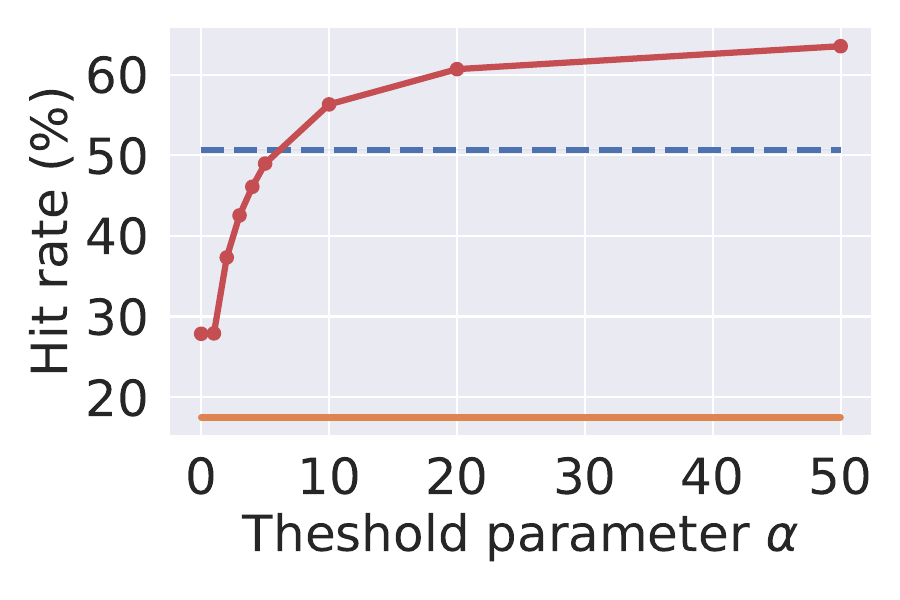} \\
	\end{tabular}
	\vspace{-5pt}
	\caption{Triplet hit rates (\S~\ref{sec:sg_quality}) versus the threshold $\alpha$ on four different VG test subsets using our perturbation methods.
	}
	\vspace{-5pt}
	\label{fig:hit_rates}
\end{figure*}

\textbf{Baselines.}
In addition to the IMP++ and NM++ baselines, we evaluate \textsc{Resample}, \textsc{Reweight} and TDE~\cite{tang2020unbiased} when combined with IMP++.
\textsc{Resample} samples training images based on the inverse frequency of predicates/triplets~\cite{tang2020unbiased}. \textsc{Reweight} increases the softmax scores of rare predicate classes (see \S~\ref{sec:reweight} in \suppl~for details).
{TDE} debiases contextual edge features of a SGG model. We use the Total Effect (TE) variant according to Eq.~6 in~\cite{tang2020unbiased}, since applying TDE to IMP++ is not straightforward due to the absence of conditioning on node labels when making predictions for edges in IMP++.
\textsc{Reweight} and TDE/TE do not require retraining IMP++.\looseness-1

\vspace{-2pt}
\textbf{GAN.} To train the generator $G$ and discriminators $D$ of a GAN, we generally follow hyperparameters suggested by SPADE~\cite{SPADE}. In particular, we use Spectral Norm~\cite{miyato2018spectral} for $D$,  Batch Norm~\cite{ioffe2015batch} for $G$, and TTUR~\cite{heusel2017gans} with learning rates of 1e-4 and 2e-4 for $G$ and $D$ respectively.

\textbf{Perturbation methods (\S~\ref{sec:perturb}).} 
We found that perturbing $L=20\%$ nodes works well across the methods, which we use in all our experiments. For \textsc{Neigh} we use top-k=10 as a compromise between too limited diversity and plausibility. For \structn, we set top-k=5, as the method enables larger diversity even with very small top-k. To train the GAN-based models with \structn, we use frequency threshold $\alpha=[2, 5, 10, 20]$.
In addition to the proposed perturbation methods, we also consider so called \oracle~perturbations.
These are created by directly using ZS triplets from the test set (all obtained triplets are the same as ZS triplets, so that zero-shot hit rate is 100\%). We also evaluate \oracle+$\hat{B}$, which in addition to exploiting test ZS triplets, uses bounding boxes from the test samples corresponding to the resulted ZS triplets. \oracle-based results are an upper bound estimate of ZS recall, highlighting the challenging nature of the task.\looseness-1

\textbf{Evaluation.} Following prior work~\cite{xu2017scene,zellers2018neural,knyazev2020graph,tang2020unbiased}, we focus our evaluation on two standard SGG tasks: scene graph classification (\textbf{SGCls}) and predicate classification (\textbf{PredCls}), using recall (R@K) metrics.
The scene graph generation (\textbf{SGGen}) results are presented in \S~\ref{sec:sggen} in \suppl.
Unless otherwise stated, we report results with K=100 for SGCls and K=50 for PredCls, since the latter is an easier task with saturated results for K=100. We compute recall \textit{without} the graph constraint in Table~\ref{table:main_results}, since it is a less noisy metric~\cite{knyazev2020graph}.
We emphasize performance metrics that focus on the ability to recognize rare and novel visual relationship compositions~\cite{knyazev2020graph,tang2020unbiased,suhail2021energy}: \textbf{zero-shot} and \textbf{10-shot} recalls.
In Tables~\ref{table:main_results} and~\ref{table:zs_results}, the mean and standard deviations of 3 runs (random seeds) are reported.

\vspace{-3pt}
\subsection{Results\label{sec:results}}
\vspace{-3pt}

\textbf{Main SGG results (Table~\ref{table:main_results}).}
First, we compare the baseline IMP++ to our GAN-based model trained \textit{without} and \textit{with} perturbation methods.
Even without any perturbations, the GAN-based model significantly outperforms IMP++, especially on the 100-shot and all-shot recalls.
GANs with simple perturbation strategies, \textsc{Rand} (as in~\cite{wang2019generating}) and \textsc{Neigh}, improve on zero-shots, but at a drop in the 100-shot and all-shot recalls. 
GANs with \structn~further improve ZS and 10-shot recalls, but compared to \textsc{Rand} and \textsc{Neigh}, also show high recalls on the 100-shots and all-shots.

For \structn, there is a connection between the SGG recall results (Table~\ref{table:main_results}) and triplet hit rates (Fig.~\ref{fig:hit_rates}) for different values of the threshold $\alpha$.
Specifically, 
\structn~with lower $\alpha$ values upsamples more of the rare compositions leading to higher ZS and 10-shot \textit{hit rate} (Fig.~\ref{fig:hit_rates} a,b) and, as a result, higher ZS and 10-shot \textit{recalls} (Table~\ref{table:main_results}).
\structn~with higher $\alpha$ values upsamples more of the frequent compositions leading to higher 100-shot and all-shot \textit{hit rates} (Fig.~\ref{fig:hit_rates}~c,d) and, as a result, higher 100-shot and all-shot \textit{recalls}.
Compared to \textsc{Rand} and \textsc{Neigh}, the compositions obtained using \structn~have higher triplet hit rates due to better respecting the graph structure and dataset statistics. As a result, \structn~shows overall better recalls in SGG, even approaching the \oracle~model (Table~\ref{table:main_results}). 
Devising a perturbation strategy universally strong across all metrics is challenging. \textsc{Neigh} can be viewed as such an attempt, which shows average hit rates for all test subsets, but lower performance in all SGG metrics.\looseness-1

\begin{table}[t]
	\footnotesize
	\begin{center}
		\caption{ZS recall results on VG using the graph constraint evaluation. $^\dagger$The results are obtained with a more advanced feature extractor and, thus, are not directly comparable.}
		\vspace{1pt}
		\setlength{\tabcolsep}{2pt}
		\label{table:zs_results}
		\begin{tabular}{lcccc}
			\multirow{2}{*}{\textsc{\textbf{Model}}} &
			\multicolumn{2}{c}{{\textbf{SGCls}}} & 
			\multicolumn{2}{c}{{\textbf{PredCls}}} \\
            & \scriptsize zsR@50 & \scriptsize zsR@100 & \scriptsize zsR@50 & \scriptsize zsR@100 \\
			\toprule
			\textsc{Freq}~\cite{zellers2018neural} & 0.0 & 0.0 & 0.1 & 0.1 \\
			KERN~\cite{chen2019knowledge} & $-$ & 1.5 & 3.9 & $-$\\
			VCTree$^\dagger$~\cite{tang2020unbiased} & 1.9 & 2.6 & 10.8 & 14.3\Bstrut\\
			\hline
			NM~\cite{zellers2018neural}  & 1.1 & 1.7 & 6.5 & 9.5\Tstrut\\
			
			NM$^\dagger$~\cite{tang2020unbiased} & 2.2 & 3.0 & 10.9 & 14.5\\
			
			NM, TDE$^\dagger$~\cite{tang2020unbiased} & 3.4 & \textbf{4.5} & 14.4 & 18.2\\
			
			NM, EBM$^\dagger$~\cite{suhail2021energy} & 1.3 & $-$ & 4.9 & $-$\\
			
			NM++~\cite{knyazev2020graph} & 1.8\std{0.1} & 2.3\std{0.1} & 10.2\std{0.1} & 13.4\std{0.3}\\
			
			NM++, GAN+\structn & 2.5\std{0.1} & 3.1\std{0.1} & 14.2\std{0.0} & 17.4\std{0.3}\Bstrut\\
			
			\hline
			
			IMP+~\cite{xu2017scene,zellers2018neural} & 2.5 & 3.2 & 14.5 & 17.2\Tstrut\\
			
			IMP+, EBM$^\dagger$~\cite{suhail2021energy} & 3.7 & $-$ & 18.6 & $-$\\
			
			IMP++~\cite{knyazev2020graph} & 3.5\std{0.1} & 4.2\std{0.2} & 18.3\std{0.4} & 21.2\std{0.5}\\
            
            IMP++, TDE & 3.5\std{0.1} & 4.3\std{0.1} & 18.5\std{0.3} & 21.5\std{0.3}\\
			
            IMP++, GAN+\structn & 3.7\std{0.1} &  4.4\std{0.1} & 19.1\std{0.3} &  21.8\std{0.4}\\
			IMP++, GAN+\structn~(max) & \textbf{3.8}  &  \textbf{4.5} & \textbf{19.5} &  \textbf{22.4}\\
			\bottomrule
		\end{tabular}
	\end{center}
	\vspace{-15pt}
\end{table}

Among the alternatives to our GAN approach, \textsc{Reweight} improves on zero-shots, 10-shots and mean recall (SGCls-mR) (Table~\ref{table:main_results}). However it downweights the class scores of frequent predicates, which directly degrades 100-shot and all-shot recalls.
\textsc{Resample} underperforms on all metrics except for SGCls-mR. The main limitation of \textsc{Resample} is that when we resample images with rare predicates/triplets, those images are likely to contain annotations of frequent predicates/triplets. 
Another method, TDE~\cite{tang2020unbiased}, only debiases the predicates similarly to \textsc{Reweight} and \textsc{Resample}-predicates. So, it may benefit little in recognizing ZS triplets such as $(cup, on, surfboard)$, because the predicate `on' is the frequent one. 
ZS compositions with such frequent predicates are abundant in VG (Fig.~\ref{fig:motivation}). Thus, debiasing only the predicates fundamentally limits TDE's performance. In contrast, our GAN method does not suffer from this limitation, since we perturb scene graphs aiming to increase \textit{compositional diversity}, not merely the frequency of rare predicates.
As a result, our GAN method improves on all metrics, \textit{especially} on ZS (in relative terms).\looseness-1

\textbf{Comparison to other SGG works (Table~\ref{table:zs_results}).}
Our GAN approach also improves ZS recall (zsR) of other SGG models, namely NM++. For example in PredCls, GAN+\structn~improves zsR of NM++ by 4 percentage points.
Compared to the other previous methods presented in Table~\ref{table:zs_results}, we obtain competitive ZS results on par or better with TDE~\cite{tang2020unbiased} and recent EBM~\cite{suhail2021energy}. However, it is hard to directly compare to the results reported in~\cite{tang2020unbiased,suhail2021energy} due to the different object detectors and potential implementation discrepancies.

\begin{table}[t]
	\centering
	\caption{Evaluation of generated (fake) node feature using the metrics of ``similarity'' between two distributions $X$ and $Y$~\cite{kynkaanniemi2019improved,naeem2020reliable}. The same held-out set of real test features ($Y \sim V$) is used as the reference distribution in all cases. The percentage in the superscripts denotes a relative drop of the average metric when switching from test to test-zs conditioning. For all metrics, higher is better.\looseness-1}
	\label{tab:gen}
	\vspace{2pt}
	\scriptsize
	\setlength{\tabcolsep}{1.5pt}
	\begin{tabular}{l|cc|cc|p{1.2cm}}
	\toprule
		\multirow{2}{*}{\tiny\bf\textsc{Distribution $X$}} & \multicolumn{2}{c|}{\bf Fidelity (realism)} & \multicolumn{2}{c|}{\bf Diversity} & \multicolumn{1}{c}{\multirow{2}{*}{\bf \textsc{Avg}}} \\
		& \bf \textsc{Precision} & \bf \textsc{Density} 
		& \bf \textsc{Recall} & \bf \textsc{Coverage} & \Bstrut\\
		\hline
		Real test  & 0.74 & 1.02 & 0.75 & 0.97 & 0.87 \Tstrut\\
		Real test-zs & 0.66 & 0.99 & 0.70 & 0.94 & 0.82$^{-6\%}$ \\
		GAN: Fake test & 0.55 & 0.77 & 0.42 & 0.82 & 0.64 \\
		GAN: Fake test-zs & 0.47 & 0.60 & 0.41 & 0.75 & 0.56$^{-13\%}$\\
	    \bottomrule
    \end{tabular}
    \vspace{-10pt}
\end{table}

\begin{figure}[t]
\centering
\footnotesize
\setlength{\tabcolsep}{1.5pt}
\begin{tabular}{cc}
      \textbf{\textsc{Real node features}} $V$ & \textbf{\textsc{Fake node features}} $\hat{V}$\vspace{1pt}\\
     \includegraphics[width=0.21\textwidth]{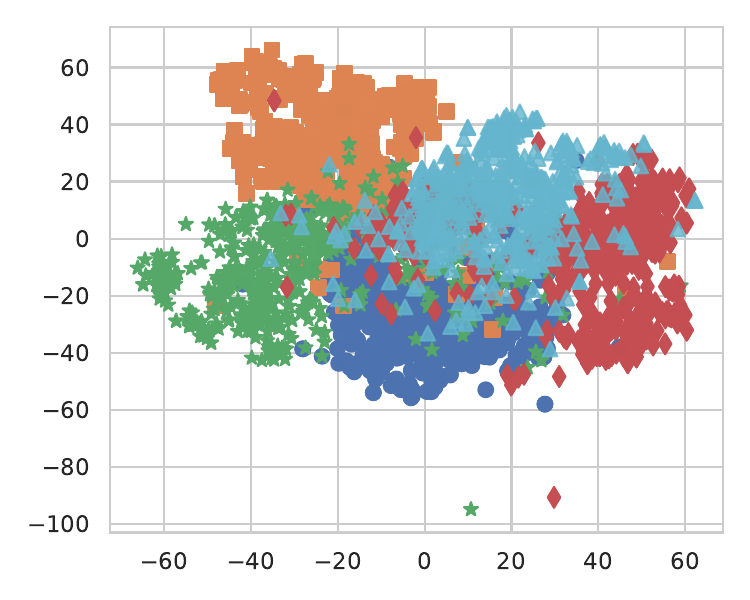} & \includegraphics[width=0.21\textwidth]{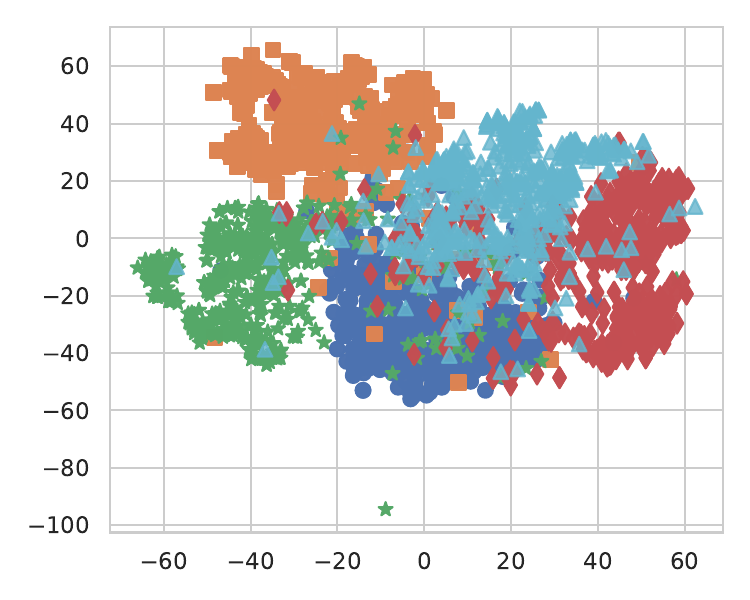}\vspace{-5pt}\\
     \multicolumn{2}{c}{{\includegraphics[width=0.46\textwidth, trim={3cm 8.5cm 0.5cm 0.6cm}, clip]{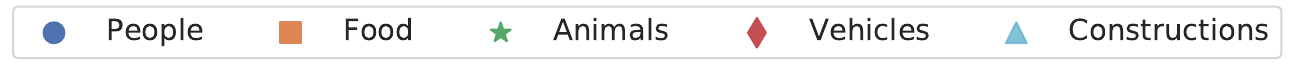}}}
\end{tabular}
  \caption{Real \textit{vs} generated node features plotted using t-SNE.}
\label{fig:tsne}
\vspace{-10pt}
\end{figure}

\textbf{Evaluation of generated visual features.}
We evaluate the quality of generated features of our GAN trained with \structn~by comparing the generated (fake) features to the real ones. To obtain fake node features $\hat{V}$, we condition our GAN on test SGs. To obtain real node features $V$, we apply the pretrained object detector to test images as described in \S~\ref{sec:generation}.
First, for the qualitative evaluation of node features, we group features based on the object category's super-type, \eg `people' includes all features of `man', `woman', `person', etc. When projected on a 2D space using t-SNE~\cite{van2008visualizing}, the fake features $\hat{V}$ generated using our GAN are clustered similarly to the real features $V$ (Fig.~\ref{fig:tsne}). Therefore, qualitatively our GAN generates realistic and diverse features given a scene graph.\looseness-1

Second, we evaluate GAN features quantitatively. For that purpose, we follow~\cite{devries2020instance} and use Precision, Recall~\cite{kynkaanniemi2019improved} and Density, Coverage~\cite{naeem2020reliable} metrics. 
These metrics compare the manifolds spanned by real and fake features and do not require any labels.
We consider two cases: conditioning our GAN on test SGs and test zero-shot (test-zs) SGs. The motivation is similar to~\cite{casanova2020generating}: understand if novel compositions confuse the GAN and lead to poor features, that in our context may result in poor training of the main model $F$. 
Indeed, the features generated conditioned on test-zs SGs significantly degrade in quality compared to test SGs, especially in terms of fidelity (Table~\ref{tab:gen}). This result suggests that it is more challenging to produce realistic features for more rare compositions limiting our approach (see \S~\ref{sec:nolimit} in \suppl~for a discussion).
The same qualitative and quantitative experiments for edge features $(E,\hat{E})$ and global features $(H,\hat{H})$ confirm our results: (1) when conditioned on test SGs, the generated features are realistic and diverse; (2) conditioning on more rare compositions degrades feature quality (see \S~\ref{apdx:gan}).\looseness-1

\begin{figure}[t]
\centering
\setlength{\tabcolsep}{0pt}
\vspace{-2pt}
\begin{tabular}{c}
\includegraphics[width=0.48\textwidth]{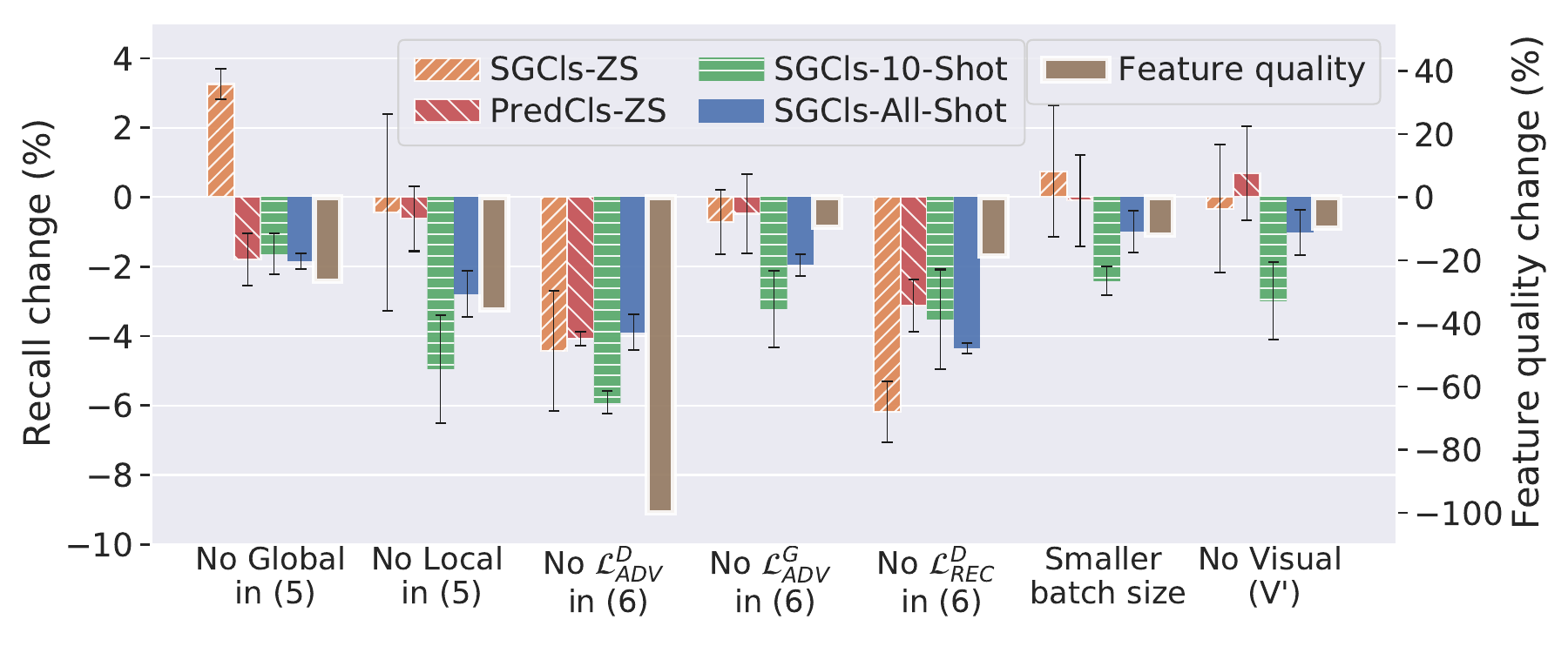} 
\end{tabular}
\vspace{-7pt}
\caption{
	Ablations of our GAN model on SGG and feature quality metrics. Error bars denote standard deviation. For feature quality the average metric on the test-zs SGs from Table~\ref{tab:gen} is used.}
\vspace{-13pt}
\label{fig:ablations}
\end{figure}

\textbf{Ablations (Figure~\ref{fig:ablations}).} We also performed ablations to determine the effect of the proposed GAN losses \eqref{eq:total_loss} and other design choices on the (i) SGG performance and (ii) quality of generated features. As a reference model, we use our GAN model without any perturbations. 
In general, all ablated GANs degrade both in (i) and (ii) with correlated drops between (i) and (ii). So by improving generative models in future work, we can expect larger SGG gains. One exception is the GAN without the global terms in \eqref{eq:adv_full}, which performed better on zero-shots despite having lower feature quality. This might be explained as some regularization effect. We also found that this model did not combine well with perturbations.

\begin{figure}[t]
    \vspace{-5pt}
	\centering
	\scriptsize
	\newcommand{\width}{0.115\textwidth}
	\setlength{\tabcolsep}{0.1pt}
	\begin{tabular}{p{0.25cm}c|c|c|c}
		& \includegraphics[width=0.11\textwidth]{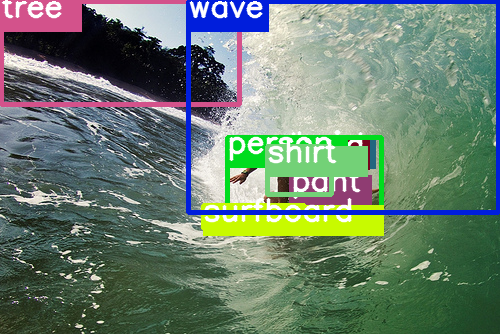} 
		& \includegraphics[width=\width,trim={2cm 0.1cm 7cm 0.1cm},clip]{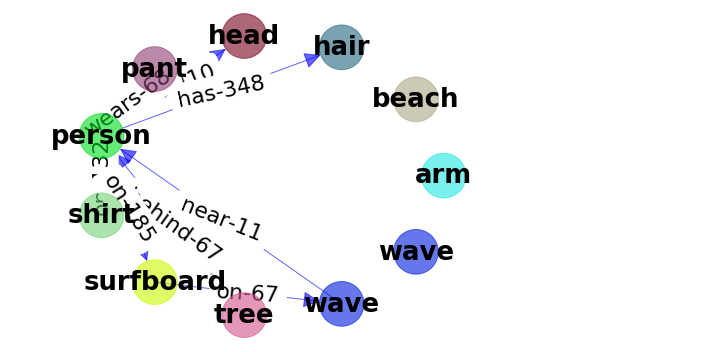} &
		\includegraphics[width=\width,trim={2cm 0.1cm 7cm 0.1cm},clip]{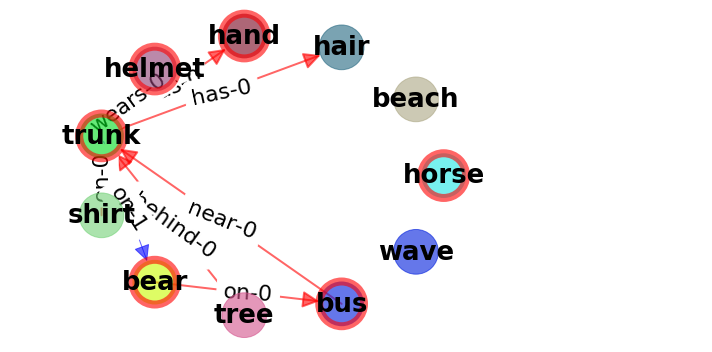} &
		\includegraphics[width=\width,trim={2cm 0.1cm 7cm 0.1cm},clip]{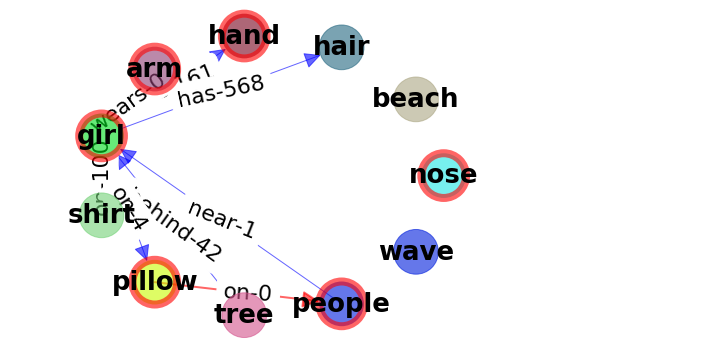} \\
		& \textsc{\textbf{Image}} & \textbf{\textsc{Original SG}} & \textbf{\textsc{Rand}} & \textbf{\textsc{Neigh}}\Bstrut\\
		\toprule 
		\multicolumn{1}{c|}{\rotatebox[origin=c]{90}{\textbf{\structn}}} & \includegraphics[align=c,width=\width,trim={2cm 0.1cm 7cm 0.1cm},clip]{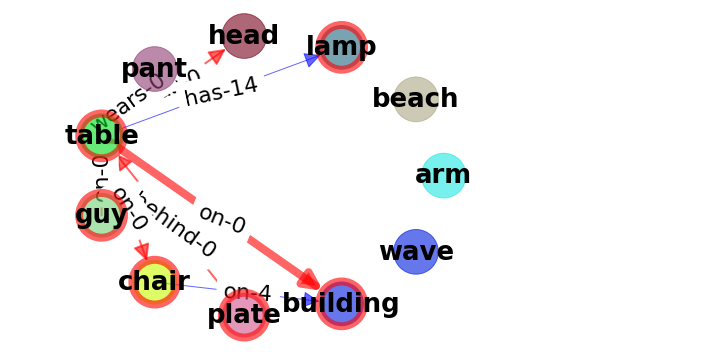} & 
		\includegraphics[align=c,width=\width,trim={2cm 0.1cm 7cm 0.1cm},clip]{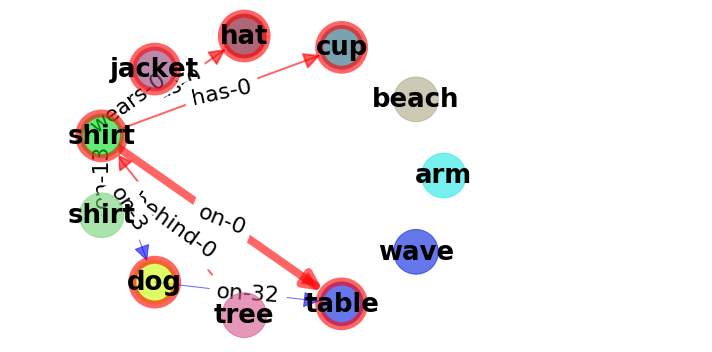} & 
		\includegraphics[align=c,width=\width,trim={2cm 0.1cm 7cm 0.1cm},clip]{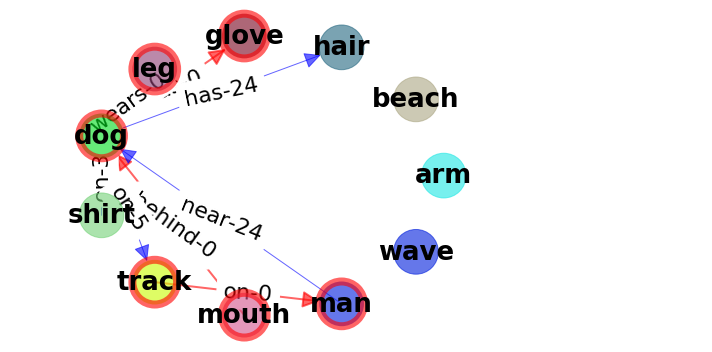} & \includegraphics[align=c,width=\width,trim={1.5cm 0.1cm 7cm 0.1cm},clip]{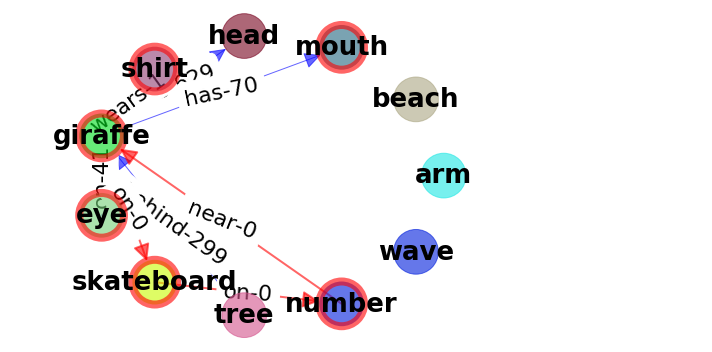}\Tstrut\\
		& $\alpha=2$ & $\alpha=5$ & $\alpha=10$ & $\alpha=20$ \\
	\end{tabular}
	\vspace{5pt}
	\caption{Examples of perturbations (nodes in red) applied to a scene graph. The numbers on edges denote the count of triplets in the training set and a thick red arrow denotes matching a ZS triplet.\looseness-1}
	\label{fig:examples}
	\vspace{-5pt}
\end{figure}

\textbf{Evaluating the quality of SG perturbations.}
We show examples of SG perturbations in Fig.~\ref{fig:examples} and in \suppl. In case of \textsc{Rand}, most of the created triplets are implausible as a result of random perturbations. \textsc{Neigh} leads to very likely compositions, but less often provides rare plausible compositions.
In contrast, \structn~can create plausible compositions that are rare or more frequent depending on $\alpha$.

We also analyzed the quality of real and perturbed SGs using the BERT-based metric (\S~\ref{sec:sg_quality}).
We found that the overall test set has on average the highest BERT scores, while lower-shot subsets gradually decrease in ``semantic plausibility'', which aligns with our intuition. We then perturbed all nodes of all test SGs using our perturbation strategies. Surprisingly, real test-zs SGs have very low plausibility close to \textsc{Rand}-based SGs. \textsc{Neigh} produces SGs of plausibility between real 10-shot and 100-shot SGs. In contrast, with \structn~we can gradually slide between low and high plausibility, which enabled better SGG results. The BERT scores, however, are not tied to the VG dataset. So, semantic plausibility per BERT may be different from the likelihood per VG.\looseness-1

\begin{figure}[t]
	\centering
	\vspace{-5pt}
	\centering
	\setlength{\tabcolsep}{0pt}
	\begin{tabular}{cc}
		\includegraphics[align=c,width=0.28\textwidth,trim={0 0 0 0},clip]{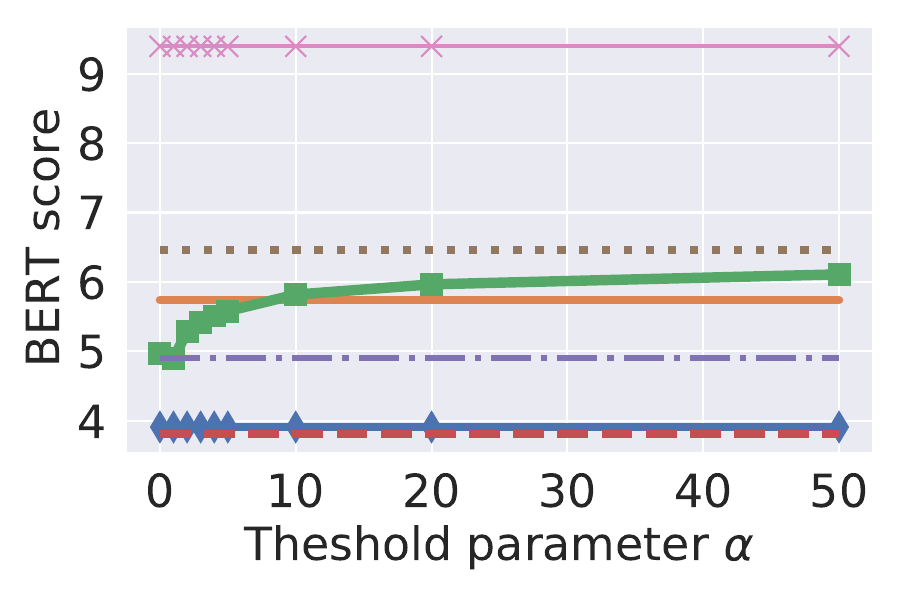} & 
        {\includegraphics[align=c,width=0.13\textwidth,trim={13.5cm 3cm 5.5cm 1.5cm},clip]{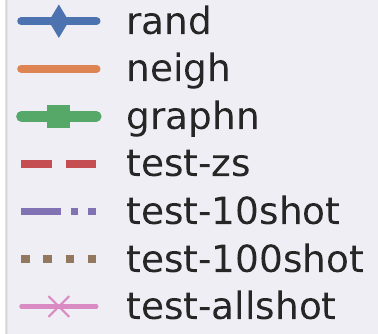}}\\
	\end{tabular}
	\vspace{1pt}
	\caption{ Semantic plausibility (as per BERT) depending on $\alpha$. These results should be interpreted with caution, because: (1) the variance of scores is very high (not shown); (2) in the zero- and few-shot test subsets the graphs are significantly smaller, which affects the amount of contextual information available to BERT. }
	\label{fig:results_semantic}
\end{figure}

\vspace{-3pt}
\section{Conclusion}
\vspace{-5pt}
We focus on the compositional generalization problem within the scene graph generation task. Our GAN-based augmentation approach can be used with different SGG models and can improve their zero-, few- and all-shot SGG results. To obtain better SGG results using our augmentations, it is important to rely on the structure of scene graphs and tune the augmentation parameters towards a specific SGG metric. Our evaluation confirmed that our augmentations provide plausible compositions and the generator generally produces high-fidelity and diverse features enabling gains in SGG.\looseness-1

\vspace{-2pt}	
\section*{\small Acknowledgments}
\vspace{-5pt}
\small
	We thank all the reviewers for their useful feedback.
	This research was developed with funding from DARPA. The views, opinions and/or findings expressed are those of the authors and should not be interpreted as representing the official views or policies of the Department of Defense or the U.S.~Government. The authors also acknowledge support from the Canadian Institute for Advanced Research and the Canada Foundation for Innovation.
	Resources used in preparing this research were provided, in part, by the Province of Ontario, the Government of Canada through CIFAR, and companies sponsoring the Vector Institute: \url{http://www.vectorinstitute.ai/#partners}.

\normalsize
\medskip

{\small
\begin{spacing}{0.95}
    \setlength{\itemsep}{0pt}
	\bibliographystyle{ieee_fullname}
	\bibliography{ref}
\end{spacing}	
}

\newpage

\appendix
\section*{Appendix}

	\section{Experimental Details\label{sec:details}}
	
	\subsection{Architectures\label{apdx:arch}}
	For our GAN model see the detailed architectures of the generator and discriminators in Tables~\ref{tab:generator_arch} and ~\ref{tab:discr_arch} respectively.
	
	\begin{table}[tbhp]
		\setlength{\tabcolsep}{0pt}
		\footnotesize
		\begin{center}
			\caption{\textbf{Architecture of the generator $G$.} All GraphTripleConv and Upsample-Refine blocks include Batch Norm (BN)~\cite{ioffe2015batch} followed by ReLU/LeakyReLU. $\tilde{n}, \tilde{m}$ are the number of nodes and edges in a batch; $bs$ is batch size.}
			\label{tab:generator_arch}
 			\vspace{2pt}
			\begin{tabular}{lcc}
				\textbf{Layer} & \textbf{Nodes} & \textbf{Edges}\\
				\toprule
				\textbf{input-1} (embedd. layer) & $\tilde{n}$x200 & $\tilde{m}$x200\\
				concat. with boxes $B$ & $\tilde{n}$x204 & $\tilde{m}$x200\\
				GraphTripleConv-1 & 204x64 & 608x64\\
				GraphTripleConv-2 & 64x64 & 192x64 \\
				GraphTripleConv-3 & 64x64 & 192x64 \\
				GraphTripleConv-4 & 64x64 & 192x64 \\
				GraphTripleConv-5 & 64x(32$\cdot$7$\cdot$7) & 192x64 \\
				\textbf{output-1} & $\tilde{n}$x32x7x7 & $-$\\
				\hline 
				conv1 & 32x64\Tstrut\\
				conv2 & 64x64 \\
				\textbf{output-2} & $\tilde{n}$x64x7x7 & \\
				\textbf{input-2}: vis. features $V^\prime$ & $\tilde{n}$x512x7x7\\ 
				concatenate & output-2, input-2 \\
				conv3 & (64+512)x64 \\
				\textbf{output-3} & $\tilde{n}$x64x7x7 &\\
				\hline 
				\textbf{input-3}: boxes $B$ & $\tilde{n}$x4\Tstrut\\
				Boxes2Layout & \multicolumn{2}{c}{build feature maps based on output-3 and $B$}\\
				\textbf{output-4} & \multicolumn{2}{c}{$bs$x64x37x37} \\
				\hline
				& \multicolumn{2}{c}{\textbf{Global feature maps}}\Tstrut\\
				Upsample-Refine-1 & \multicolumn{2}{c}{64x128}\\
				Upsample-Refine-2 & \multicolumn{2}{c}{128x256} \\
				Upsample-Refine-3 & \multicolumn{2}{c}{256x512} \\
				conv1x1 & \multicolumn{2}{c}{512x512} \\
				\textbf{output-5} ($\hat{H}$) & \multicolumn{2}{c}{$bs$x512x37x37} \\
				\hline
				RoIAlign & \multicolumn{2}{c}{extract node, edge feat. based on output-5 and $B$}\Tstrut\\
				\textbf{output-6} ($\hat{V}, \hat{E}$) & $\tilde{n}$x512x7x7 & $\tilde{m}$x512x7x7\\
				\bottomrule
			\end{tabular}
		\end{center}
		\vspace{-15pt}
	\end{table}
	
	\begin{table}[tbhp]
		\setlength{\tabcolsep}{2pt}
		\footnotesize
		\begin{center}
			\caption{\textbf{Architectures of the discriminators $D$.} Following~\cite{radford2015unsupervised}, the discriminators are regularized by having a fully-convolutional architecture with four layers interleaved with ReLU. All convolutions are 3x3 without padding and normalized with Spectral Norm (SN)~\cite{miyato2018spectral}. $\tilde{n}, \tilde{m}$ are the number of nodes and edges in a batch; $bs$ is batch size;  
				$|{\cal C}|=151, |{\cal R}|=51$ - number of object and predicate classes including the ``background'' (no object, no edge) class~\cite{zellers2018neural}.}
			\label{tab:discr_arch}
			\begin{tabular}{lccc}
				\textbf{Layer} & $D_{\text{node}}$ &  $D_{\text{edge}}$ & $D_{\text{global}}$ \\
				\toprule
				\textbf{input} & $\tilde{n}$x(512+151)x7x7 & $\tilde{m}$x(512+51)x7x7 & $bs$x512x37x37\Tstrut\\
				real feature/labels & $V,O$ & $E,R$ & $H$\Tstrut\\
				fake feature/labels & $\hat{V},\hat{O}$ & $\hat{E},R$ & $\hat{H}$ \\
				\hline
				SN-conv1 & (512+151)x256 & (512+51)x256 & 512x256\Tstrut\\
				nonlinearity & ReLU & ReLU & LeakyReLU \\
				pooling & $-$ & $-$ & Average-2 \\
				SN-conv2 & 256x128 & 256x128 & 256x256\Tstrut\\
				nonlinearity & ReLU & ReLU & LeakyReLU \\
				pooling & $-$ & $-$ & Average-2 \\
				SN-conv3 & 128x64 & 128x64 & 256x128\Tstrut\\
				nonlinearity & ReLU & ReLU & LeakyReLU \\
				pooling & $-$ & $-$ & Average-2 \\
				SN-conv4 & 64x1 & 64x1 & 128x1\Tstrut\\
				\hline
				\textbf{output} & $\tilde{n}$x1 & $\tilde{m}$x1 & $bs$x1\Tstrut\\
				\bottomrule
			\end{tabular}
		\end{center}
		\vspace{-20pt}
	\end{table}

	In the generator, GraphTripleConvNet\footnote{GraphTripleConvNet: \url{https://github.com/google/sg2im/blob/master/sg2im/graph.py}},
	Boxes2Layout\footnote{Boxes2Layout: \url{https://github.com/google/sg2im/blob/master/sg2im/layout.py}},
	Upsample-Refine\footnote{Upsample-Refine: \url{https://github.com/google/sg2im/blob/master/sg2im/crn.py}} are borrowed from the sg2im implementation of~\cite{johnson2018image}.
	For the baseline IMP++ model we use a publicly available implementation\footnote{IMP+/IMP++: \url{https://github.com/rowanz/neural-motifs}} with a default architecture and the graph-normalized loss~\cite{knyazev2020graph} from another public code\footnote{Graph-normalized loss: \url{https://github.com/bknyaz/sgg}}.

	\begin{table*}[t!]
		\setlength{\tabcolsep}{4pt}
		\scriptsize
		
		\begin{center}
			\caption{Results of \structn~with $\text{topk}=5$ (same as reported in the main text) compared to \textsc{Neigh} with different values of topk. The models are based on IMP++~\cite{zellers2018neural,knyazev2020graph}.}
			\vspace{3pt}
			\label{tab:topk_results}
			\begin{tabular}{p{3.5cm}ccp{0.1cm}ccp{0.1cm}ccp{0.1cm}ccc}
				\multirow{2}{*}{\textsc{\textbf{Model}}} & \multicolumn{2}{c}{\textsc{\textbf{Zero-shot Recall}}} & &
				\multicolumn{2}{c}{\textsc{\textbf{10-shot Recall}}} & &
				\multicolumn{2}{c}{\textsc{\textbf{100-shot Recall}}} & & 
				\multicolumn{3}{c}{\textsc{\textbf{All-Shot Recall}}} \\
				 &
				\multicolumn{1}{c}{\scriptsize{SGCls}} & \multicolumn{1}{c}{\scriptsize{PredCls}} & &
				\multicolumn{1}{c}{\scriptsize{SGCls}} & \multicolumn{1}{c}{\scriptsize{PredCls}} & & \multicolumn{1}{c}{\scriptsize{SGCls}} & \multicolumn{1}{c}{\scriptsize{PredCls}} & & \multicolumn{1}{c}{\scriptsize{SGCls}} & \multicolumn{1}{c}{\scriptsize{PredCls}} & \multicolumn{1}{c}{\scriptsize{SGCls-mR}}\\
				\cline{2-3}\cline{5-6}
				\toprule
				 
			GAN+\structn, $\alpha=2$ & \textbf{9.89}\std{0.15} & 28.90\std{0.14} & & 21.96\std{0.30} & \textbf{43.79}\std{0.27} & & 41.22\std{0.33} & 69.17\std{0.24} & & 50.06\std{0.29} & 78.98\std{0.09} & 27.79\std{0.48}\\
			
			GAN+\structn, $\alpha=5$ & 9.62\std{0.29} & \textbf{29.18}\std{0.33} & & \textbf{22.24}\std{0.11} & 43.74\std{0.10} & & 41.39\std{0.26} & 69.11\std{0.05} & & \textbf{50.14}\std{0.21} & 78.94\std{0.03} & \textbf{27.98}\std{0.23}\\
			
			GAN+\structn, $\alpha=10$ & 9.84\std{0.17} & 28.90\std{0.46} & & 22.04\std{0.33} & 43.54\std{0.36} & & \textbf{41.46}\std{0.15} & 69.13\std{0.24} & & 50.10\std{0.23} & \textbf{79.00}\std{0.09} & 27.68\std{0.37}\\
			
			GAN+\structn, $\alpha=20$ & 9.65\std{0.15} & 28.68\std{0.28} & & 21.97\std{0.30} & 43.64\std{0.20} & & 41.24\std{0.08} & \textbf{69.31}\std{0.17} & & 49.89\std{0.28} & 78.95\std{0.04} & 27.42\std{0.36}\Bstrut\\ 
			
				\hline\hline
				
				GAN+\textsc{Neigh}, $\text{topk}=2$ & 9.49\std{0.21} & 28.58\std{0.40} & & 21.72\std{0.23} & 43.62\std{0.05} & & 41.04\std{0.26} & 69.07\std{0.09} & & 49.64\std{0.29} & 78.94\std{0.11} & 27.33\std{0.41}\Tstrut\\
				
				GAN+\textsc{Neigh}, $\text{topk}=5$ & 
				9.38\std{0.25} & 28.68\std{0.40} & & 21.75\std{0.25} & 43.45\std{0.14} & & 41.05\std{0.49} & 68.94\std{0.14} & & 49.75\std{0.30} & 78.94\std{0.10} & 27.05\std{0.15}\\
				
				GAN+\textsc{Neigh}, $\text{topk}=10$ & 
				9.58\std{0.22} & 28.63\std{0.39} & & 21.86\std{0.23} & 43.77\std{0.15} & & 41.14\std{0.30} & 69.03\std{0.09} & & 49.86\std{0.38} & 78.89\std{0.01} & 27.41\std{0.51}\\
				
				GAN+\textsc{Neigh}, $\text{topk}=20$ & 9.65\std{0.04} & 28.57\std{0.06} & & 21.82\std{0.17} & 43.28\std{0.21} & & 40.98\std{0.30} & 69.01\std{0.14} & & 49.68\std{0.28} & 78.92\std{0.05} & 27.17\std{0.12}\Bstrut\\

		\end{tabular}
	\end{center}
	\vspace{-10pt}
	\end{table*}
		
	\begin{table*}[h!]
		\vspace{-1pt}
		\setlength{\tabcolsep}{4pt}
		\scriptsize
		\begin{center}
			\caption{Results using models based on Neural Motifs++~\cite{zellers2018neural,knyazev2020graph}.}
			\vspace{3pt}
			\label{tab:nm_results}
			\begin{tabular}{p{3.5cm}ccp{0.1cm}ccp{0.1cm}ccp{0.1cm}ccc}
				\multirow{2}{*}{\textsc{\textbf{Model}}} & \multicolumn{2}{c}{\textsc{\textbf{Zero-shot Recall}}} & &
				\multicolumn{2}{c}{\textsc{\textbf{10-shot Recall}}} & &
				\multicolumn{2}{c}{\textsc{\textbf{100-shot Recall}}} & & 
				\multicolumn{3}{c}{\textsc{\textbf{All-Shot Recall}}} \\
				&
				\multicolumn{1}{c}{\scriptsize{SGCls}} & \multicolumn{1}{c}{\scriptsize{PredCls}} & &
				\multicolumn{1}{c}{\scriptsize{SGCls}} & \multicolumn{1}{c}{\scriptsize{PredCls}} & & \multicolumn{1}{c}{\scriptsize{SGCls}} & \multicolumn{1}{c}{\scriptsize{PredCls}} & & \multicolumn{1}{c}{\scriptsize{SGCls}} & \multicolumn{1}{c}{\scriptsize{PredCls}} & \multicolumn{1}{c}{\scriptsize{SGCls-mR}}\\
				\cline{2-3}\cline{5-6}
				\toprule
				Neural Motifs++ & 6.81\std{0.10} & 16.91\std{0.31} & & 21.07\std{0.03} & 44.75\std{0.12} & & 40.16\std{0.10} & 73.58\std{0.14} & & 48.30\std{0.06} & 82.17\std{0.09} & 25.48\std{0.15}\\
                
                Neural Motifs++, GAN+\structn & \textbf{8.06}\std{0.13} & \textbf{23.36}\std{0.08} & & 20.99\std{0.04} & \textbf{45.98}\std{0.04} & & 39.92\std{0.05} & \textbf{73.97}\std{0.13} & & 48.05\std{0.12} & \textbf{82.89}\std{0.08} & \textbf{25.59}\std{0.02}\\
				\bottomrule
			\end{tabular}

		\end{center}
	\end{table*}
	
	\begin{table*}[h!]
	\vspace{-5pt}
		\setlength{\tabcolsep}{3pt}
		\scriptsize
		\begin{center}
			\caption{Comparison of GAN models using GT versus predicted bounding boxes and \oracle~perturbations.}
			\vspace{3pt}
			\label{tab:box_pred}
			\begin{tabular}{lccp{0.1cm}ccp{0.1cm}ccp{0.1cm}ccc}
				\multirow{2}{*}{\textsc{\textbf{Model}}} & \multicolumn{2}{c}{\textsc{\textbf{Zero-shot Recall}}} & &
				\multicolumn{2}{c}{\textsc{\textbf{10-shot Recall}}} & &
				\multicolumn{2}{c}{\textsc{\textbf{100-shot Recall}}} & & 
				\multicolumn{3}{c}{\textsc{\textbf{All-Shot Recall}}} \\
				 &
				\multicolumn{1}{c}{\scriptsize{SGCls}} & \multicolumn{1}{c}{\scriptsize{PredCls}} & &
				\multicolumn{1}{c}{\scriptsize{SGCls}} & \multicolumn{1}{c}{\scriptsize{PredCls}} & & \multicolumn{1}{c}{\scriptsize{SGCls}} & \multicolumn{1}{c}{\scriptsize{PredCls}} & & \multicolumn{1}{c}{\scriptsize{SGCls}} & \multicolumn{1}{c}{\scriptsize{PredCls}} & \multicolumn{1}{c}{\scriptsize{SGCls-mR}}\\
				\cline{2-3}\cline{5-6}
				\toprule
				
				GAN+\textsc{Oracle-ZS}, $\hat{B}=B$ & 10.11\std{0.34} & 29.27\std{0.10} & & 22.05\std{0.38} & 43.78\std{0.09} & & 41.38\std{0.50} & 69.06\std{0.16} & & 50.19\std{0.36} & 79.00\std{0.08} & 27.91\std{0.56}\Tstrut\\
				GAN+\textsc{Oracle-ZS}~$\pgraph + \text{test } \hat{B}$ & 10.52\std{0.31} & 29.43\std{0.42} & & 21.98\std{0.39} & 43.03\std{0.13} & & 41.12\std{0.19} & 68.73\std{0.17} & & 50.05\std{0.35} & 78.65\std{0.09} & 27.52\std{0.46}\\
				GAN+\textsc{Oracle-ZS} $\pgraph$, pred. $\hat{B}$ (\S~\ref{sec:pred_box}) & 9.92\std{0.13} & 28.93\std{0.63} & & 21.67\std{0.36} & 42.65\std{0.47} & & 40.96\std{0.48} & 68.23\std{0.28} & & 49.80\std{0.50} & 78.55\std{0.14} & 27.22\std{0.39}\\
				\bottomrule
			\end{tabular}
		\end{center}
		\vspace{-10pt}
	\end{table*}

	\subsection{Hyperparameters and Tasks}
	\textbf{SGCls.} We use batch size $bs=24$ and $lr=(1e-3) \times bs$. Training was done on NVIDIA V100 with 32GB of GPU memory or RTX6000 with 24GB. 
	We train all models for a fixed number of 20 epochs according to~\cite{knyazev2020graph}.
	\textbf{PredCls} results are obtained following a standard procedure~\cite{zellers2018neural} of reusing the SGCls model and setting object classes to ground truth. We use PyTorch 1.5+ to train all models. The difference between the graph constraint and no constraint metrics is discussed in~\cite{zellers2018neural,knyazev2020graph}.
	We also report the results on the {\textbf{SGGen/SGDet}} task in \S~\ref{sec:sggen}. However, we highlight that {{SGGen/SGDet}} is not the focus of our work and we do not expect large improvements, since we do not update the detector on generated features.

    \subsection{Implementing IMP with TDE}
    To apply TDE to a SGG model, the model is assumed to have a contextual layer, such as LSTM in Neural Motifs or TreeLSTM in VCTree, \textit{conditioned} on object predictions from the object detector. This contextual layer to some extent captures the frequency bias in the dataset \textit{during training}. So, the main goal of TDE is to debias this contextual layer \textit{at test time}. Since, IMP does not have such a conditional contextual layer and, consequently, is less biased, applying TDE is both (1) unclear implementation-wise and (2) has questionable benefits from the conceptual point of view. 
    It is still possible to use Total Effect (TE) according to Eq.~6 in~\cite{tang2020unbiased}, which we report in the main text. But, there is almost no gain w.r.t.~IMP in this case.\looseness=-1

	\section{Additional Experimental Results\label{sec:extra}}
	All presented results in this work are on Visual Genome~\cite{krishna2017visual} (split of~\cite{xu2017scene}).
	
	\subsection{Topk Semantic Neighbors}
	In the main text, we used $\text{topk}=5$ for \structn~and $\text{topk}=10$ for \textsc{Neigh} to allow for the same level of diversity of perturbations in both strategies. To make sure \structn~outperforms not just because of a better choice of topk, we report results of \textsc{Neigh} for more values in Table~\ref{tab:topk_results}. Overall, different values of topk do not allow \textsc{Neigh} to achieve the same level of performance as \structn.
	
	\subsection{BERT-based Evaluation\label{apdx:bert}}
	
\begin{figure}[tpbh]
    \centering
    \vspace{-5pt}
    {\includegraphics[width=0.48\textwidth,trim={0.1cm 0.1cm 0.1cm 0.1cm},clip]{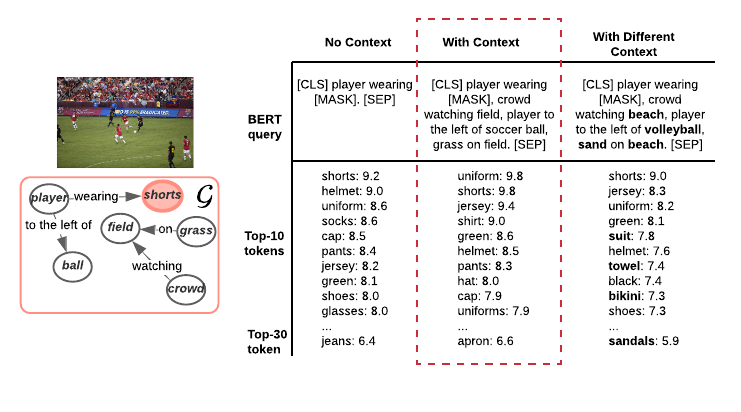}}
    \vspace{-23pt}
    \caption{Example of BERT-based evaluation. The evaluation in a dashed rectangle is the one we used in the paper. The other two options are shown for comparison. `No context' returns high scores for implausible tokens, such as `glasses'. Using the context is important for more correct evaluation, while changing the context affects the scores accordingly. The scene graph is taken from original Visual Genome~\cite{krishna2017visual} for visualization purposes. }
    \vspace{-15pt}
    \label{fig:bert_eval}
    \end{figure}

	We show an example of BERT-based evaluation in Fig.~\ref{fig:bert_eval}. In this example, we retrieve the score of 9.8 for the token `shorts' which was masked out in the query. 
	To estimate the overall likelihood of the scene graph we mask out only one node per graph for faster evaluation, which can explain a high variance of SG quality. Aggregating the scores for all nodes should lead to better estimates.
	
	\subsection{Learning to Predict Bounding Boxes\label{sec:pred_box}}
\begin{figure}[t]
		\centering
		\vspace{-5pt}
		\centering
		\begin{tabular}{cc}
			\multicolumn{2}{c}{\includegraphics[width=0.45\textwidth,trim={0 0 0 0},clip]{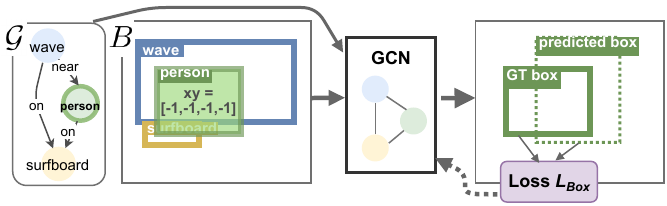}} \\
		\end{tabular}
		\caption{Overview of our model learning to predict bounding boxes, which is used only to report results in the last row of Table~\ref{tab:box_pred}.}
		\vspace{-10pt}
		\label{fig:boxgen}
	\end{figure}

	In the main text, in our generative pipeline for simplicity we assumed the layout (bounding boxes $B$) is unchanged after perturbing the associated scene graph: $\hat{B}=B$. 
	Here, we describe and report results of the alternative version, where the boxes $\hat{B}$ are predicted by some model $f$ given a perturbed scene graph $\hat{B}=f(\pgraph)$. We borrow the same principle as we used for producing $\pgraph$: instead of producing $\hat{B}$ from scratch (without relying on $B$), we perturb ground truth $B$, so $\hat{B}=f(\pgraph, B)$. This is an easier task, since we only need to predict bounding boxes for part of the scene graph.
	To achieve that, we train a simple graph convolution network (GCN) to predict a single bounding box $\hat{b}_i$ given the rest of the layout similar to autoregressive models: $\hat{b}_i=f(\graph, B, i)$, where the input coordinates of the $i$-th object in $B$ have values $b_i=[-1, -1, -1, -1]$ (Fig.~\ref{fig:boxgen}). Our $f$ is similar to the GCN in our GAN pipeline and is based on GraphTripleConv, but has 3 layers with 64 hidden units and an additional MLP which predicts 4 coordinates of the bounding box. The model is trained with a margin $l_1$ loss:
	${\cal L}_{\text{box}}=\text{min}(0.5S, \text{max}(0.05S, |\hat{b}_i - b_i|_1))$, where $S$ is the maximum box coordinate. The idea behind this loss is to avoid having too large or too small penalties making the training more stable given that there is no single correct prediction in this task (i.e. besides the GT there can be many other plausible coordinates).

	\begin{table}[t]
		\caption{Evaluation of the bounding box prediction model. FD is the Fréchet distance (lower is better). IoU denotes percentage of times the intersection over union between the predicted and GT box is $\geq 50\%$ (higher is better).} 
		\label{tab:gen_boxes}
		\vspace{5pt}
		\centering
		\footnotesize
		\setlength{\tabcolsep}{3pt}
		\begin{tabular}{lcc}
			\toprule

			\multirow{2}{*}{\textsc{\textbf{Model}}} & \textbf{FD} & \textbf{IoU (\%)} \\
			& test/test-zs & test/test-zs \\
			\hline \\
			No GCN, unconditional (sample GT) & 0.001 / 0.001 & 1.8 / 2.0 \\
			GCN, predict $b_i$ (corrupted label in $\graph$) & 0.019 / 0.034 & 2.8 / 2.2 \\ 
			GCN, predict $b_i$ & 0.018 / 0.037 & 12.9 / 6.3 \\
			GCN, GT $b_i$ & $1e{-4}$ / $3e{-4}$ & 100 / 100 \\
			\bottomrule
		\end{tabular}
		\vspace{-10pt}
	\end{table}
	
	\begin{figure}[t]
		\centering
		\begin{tabular}{cc}
			\includegraphics[width=0.2\textwidth]{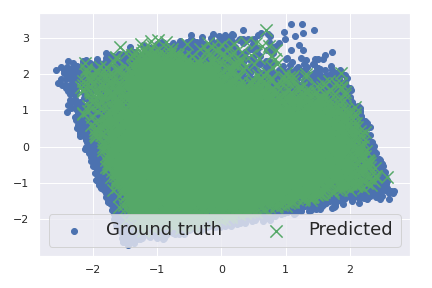} & \includegraphics[width=0.2\textwidth]{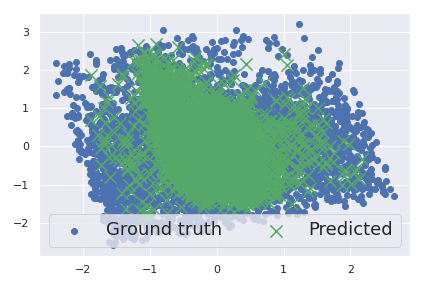}
		\end{tabular}
		\caption{PCA-projected distributions of GT and predicted bounding boxes of the test \textbf{(left)} and test zero-shot \textbf{(right)} sets.}
		\label{fig:pca_boxes}
		\vspace{-15pt}
	\end{figure}
	
	During training and for evaluation (Table~\ref{tab:gen_boxes}), we randomly (uniformly) sample a single object in a scene graph and predict its coordinates given other bounding boxes and a SG.
	After the model is trained, we sequentially apply it only to perturbed nodes keeping the boxes of non-perturbed nodes unchanged. To isolate the effect of perturbation methods, we evaluate this model using \oracle~(Table~\ref{tab:box_pred}). Surprisingly, this model performs worse on all metrics compared to both keeping the layout unchanged and using test bounding boxes.\looseness-1
	
	We estimated the quality of the bounding boxes predicted by our GCN (Table~\ref{tab:gen_boxes}). We found that the model respects the conditioning: results of IoU are significantly better compared to the case when we simply sample GT boxes from a distribution for a given class (ignoring SG and other boxes) or when we condition the model on a corrupted (random) label. However, we found that the model performs significantly worse when is fed with ZS compositions.
	In particular, the distribution appears to be more concentrated around the mean in such cases (Fig.~\ref{fig:pca_boxes}). This might explain relatively poor SGG results when we attempt to use the box prediction model in combination with \oracle~perturbations. Further research is required to resolve the challenges of reliably predicting the layout for rare compositions similarly to~\cite{casanova2020generating}.

	\subsection{Predicate Reweighting and Mean Recall\label{sec:reweight}}
	
\begin{figure}[t]
		\centering
		\setlength{\tabcolsep}{0pt}
		\begin{tabular}{cc}
			\includegraphics[width=0.23\textwidth]{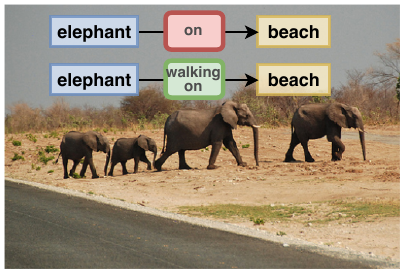} & \includegraphics[width=0.23\textwidth]{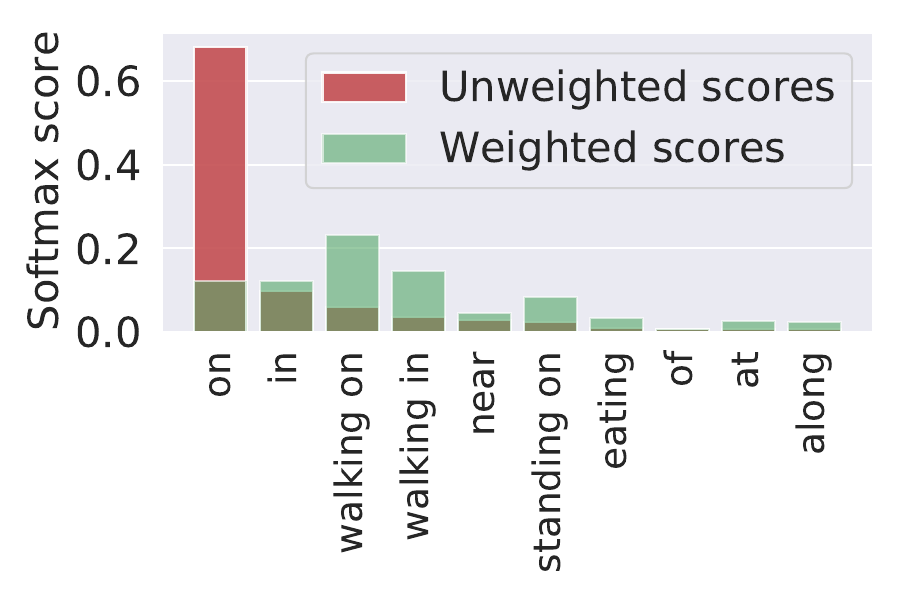} \\
		\end{tabular}
		\vspace{-1pt}
		\caption{Example of the prediction for a zero-shot triplet. Weighting predicate scores increases the score of more descriptive predicates such as 'walking on' compared to 'on', improving mR results, which aligns with~\cite{tang2020unbiased}.}
		\vspace{-10pt}
		\label{fig:weighting}
	\end{figure}

	\cite{tang2020unbiased} showed that mean (over predicates) recall is relatively easy to improve by standard Reweight/Resample methods, while ZS recall is not.
	To further analyze this and highlight the challenging nature of compositional generalization (targeted in this work) as opposed to predicate imbalance typically targeted in some previous work, we simplify and extend the Reweight idea from~\cite{tang2020unbiased}.
	In particular, we compute the frequency $f_r$ of each predicate class $r$ in the training set. Then, given a \textit{trained} SGG model, we multiply softmax scores of predicates by $w^x = 1/f_r^x$, where $x$ controls how much we need to balance predicate scores (Fig.~\ref{fig:weighting}). We consider $x=[0, 1, 2, 3]$, denoting models as $\times w^x$, where $x=0$ corresponds to not changing predictions (our default setting). 
	The advantage of our post-hoc calibration compared to Reweight~\cite{tang2020unbiased} is that it does not require retraining the model and allows us to balance performance on frequent and infrequent predicates by carefully choosing $x$.\looseness-1

	\begin{figure}[t]
		\centering
		\setlength{\tabcolsep}{2pt}
		\begin{tabular}{cc}
			\includegraphics[width=0.23\textwidth,align=c]{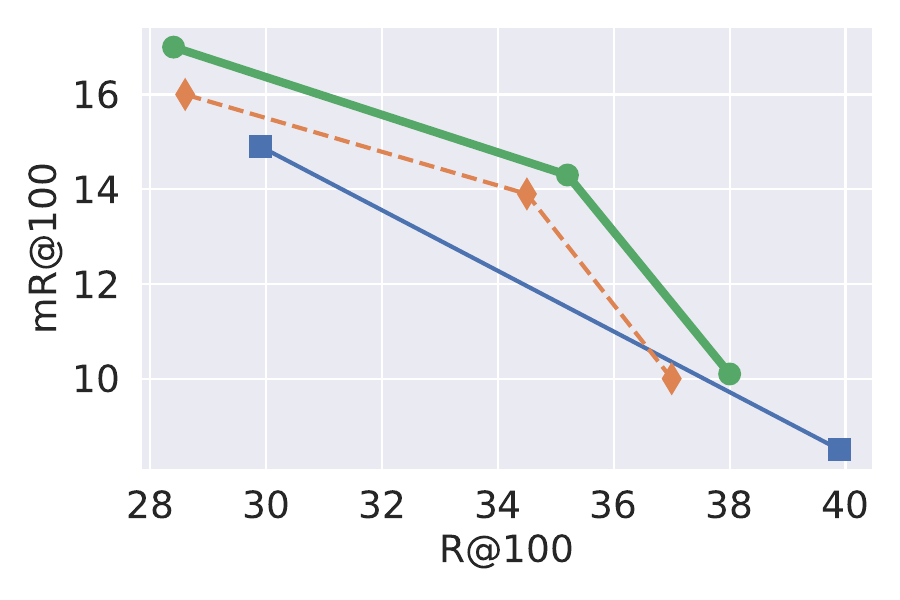} & \includegraphics[width=0.23\textwidth,align=c]{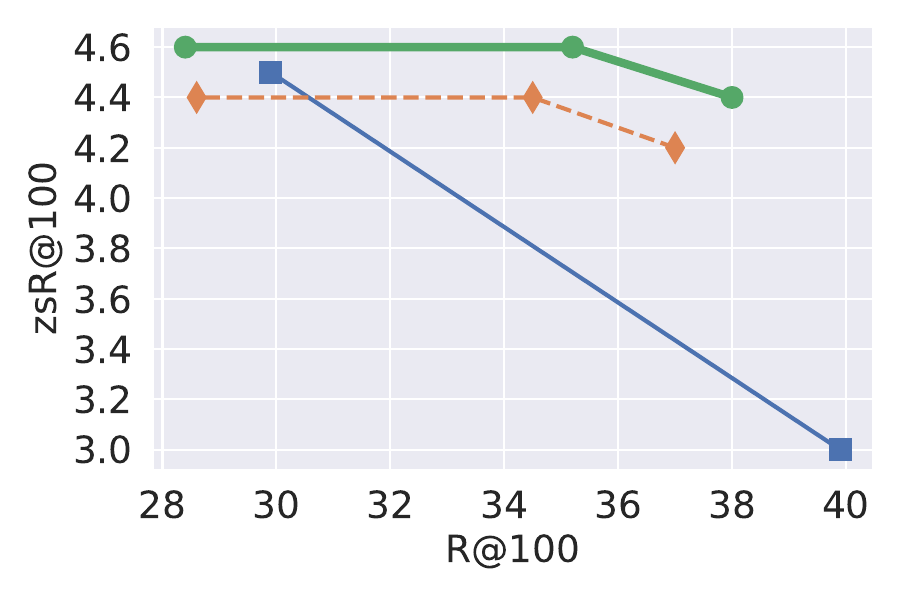} \\ \includegraphics[width=0.23\textwidth,align=c]{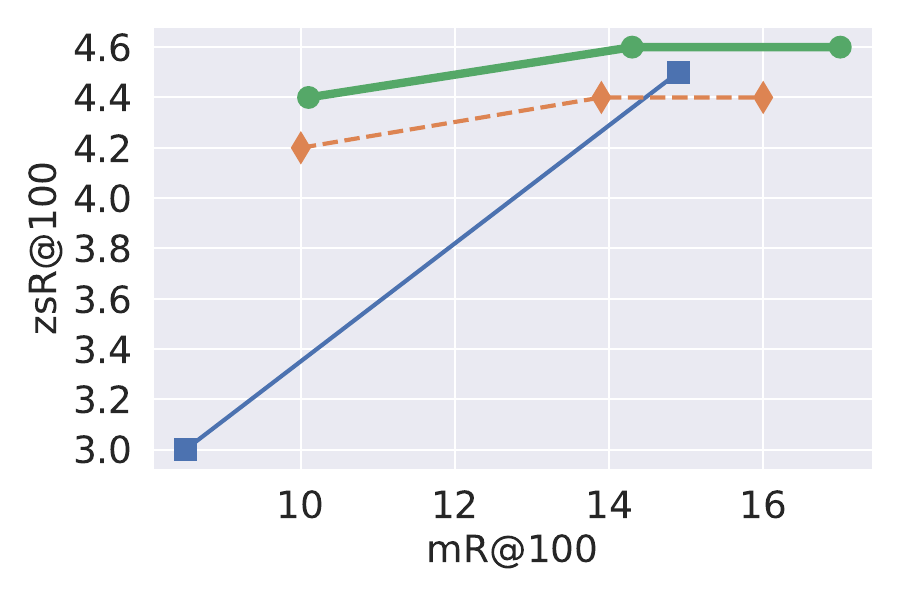} &
			{\includegraphics[width=0.2\textwidth,align=c,trim={20cm 4cm 11cm 0.5cm},clip]{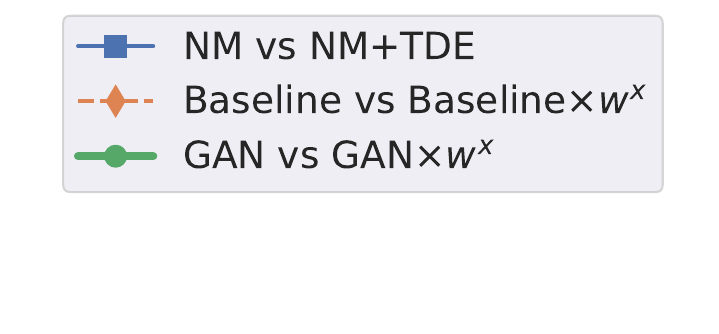}}\\	
		\end{tabular}
		\caption{Trade-off between different metrics (R, mR and zsR) in the SGCls task depending on the strength ($x=[0,1,2]$) of weighting predicate scores ($\times w^x$) or using TDE~\cite{tang2020unbiased}. }
		\vspace{-10pt}
		\label{fig:tradeoff}
	\end{figure}
	
		\begin{table}[t]
		\setlength{\tabcolsep}{3pt}
		\footnotesize
		\begin{center}
			\caption{Tuning for mean (over predicates) recall versus all-shot recall using our GAN+\structn~model (the graph constraint evaluation). Underlined are top results among the variants of our model. $^\dagger$The results in~\cite{tang2020unbiased} are obtained with a more advanced detector and, thus, are not directly comparable. }
			\vspace{5pt}
			\label{tab:mr_results}
			\begin{tabular}{lccp{0.3cm}cc}
				\multirow{2}{*}{\textsc{\textbf{Model}}} & \multicolumn{2}{c}{\textsc{\textbf{Mean Recall}}} & & \multicolumn{2}{c}{\textsc{\textbf{All-Shot Recall}}} \\
				 &
				\multicolumn{1}{c}{\scriptsize{SGCls}} & 
				\multicolumn{1}{c}{\scriptsize{PredCls}} & & \multicolumn{1}{c}{\scriptsize{SGCls}} & 
				\multicolumn{1}{c}{\scriptsize{PredCls}}\\

				\toprule
				GPS-net~\cite{lin2020gps} & 12.6 & 21.3 & & \textbf{40.1} & \textbf{66.9} \\
				NM$^\dagger$~\cite{tang2020unbiased} & 8.5 & 14.6 & & 39.9 & 66.0 \\
				NM+TDE$^\dagger$~\cite{tang2020unbiased} & 14.9 & 25.5 & & 29.9 & 46.2 \\
				PCPL~\cite{yan2020pcpl} & \textbf{19.6} & \textbf{35.2} & & 28.4 & 50.8\Bstrut\\
				\hline
				
				No weighting ($\times w^0$) & 10.1\std{0.2} & 15.7\std{0.3} & & \underline{38.1}\std{0.1} & \underline{62.3}\std{0.2}\Tstrut\\
				$\times w^1$ & 14.3\std{0.2} & 22.1\std{0.3} & & 35.2\std{0.1} & 56.9\std{0.3}\\
				$\times w^2$ & 17.0\std{0.1} & 26.3\std{0.1} & & 26.4\std{0.4} & 42.5\std{0.2}\\
				$\times w^3$ & \underline{18.3}\std{0.1} & \underline{28.6}\std{0.2} & & 17.2\std{0.1} & 27.1\std{0.2}\\
				\bottomrule
			\end{tabular}
		\end{center}
		\vspace{-15pt}
	\end{table}

	The results indicate a severe trade-off between the mean and all-shot recalls (Table~\ref{tab:mr_results}). We can achieve very competitive mean recall results by simply using large $x$. However, this dramatically hurts the all-shot recall, potentially leading to invalid predictions, such as ``cup walking on table'' instead of ``cup on table'', as discussed in~\cite{yan2020pcpl}. Overall, given the three SGG metrics (R, mR and zsR), we observe that (R, mR) and (R, zsR) are two conflicting pairs of metrics, while (mR, zsR) have some correlation (Fig.~\ref{fig:tradeoff}). Clearly, the discussed metrics measure different properties of a given SGG model. In this work, we measure compositional generalization and therefore focus on zero-shot (and closely related few-shot) recall as appropriate metrics for that particular task.

	\subsection{Training the Baseline Longer}
	
	In our GAN model we have two loss terms (${\cal L}_{\text{CLS}}, {\cal L}_{\text{REC}}$) that update the model $F$, which can be considered as having two times more updates compared to the baseline, and so can be an implicit advantage. We investigated if the baseline model can improve itself if it is updated two times more. Our results in Table~\ref{tab:updates} suggest that the model starts to overfit leading to poor results on all metrics except mean recall. This might indicate that a stronger regularization is required for the baseline trained longer. Our GAN losses can be viewed as such a regularizer leading to better generalization results.\looseness-1
	
	\begin{table}[tbph]
		\caption{Effect of training the baseline model longer. Metrics are R@100 for SGCls and R@50 for PredCls (no graph constraint).}
		\label{tab:updates}
		\vspace{1pt}
		\centering
		\footnotesize
		\setlength{\tabcolsep}{1pt}
		\begin{tabular}{lcccccccc}
			\multirow{2}{*}{\textsc{\textbf{Model}}} & \multicolumn{2}{c}{\textsc{\textbf{\scriptsize Zero-shot}}} & & \multicolumn{2}{c}{\textsc{\textbf{\scriptsize Mean Recall}}} & & \multicolumn{2}{c}{\textsc{\textbf{\scriptsize All-Shot}}} \\
			\cline{2-3}\cline{5-6}\cline{8-9}
			& \scriptsize SGCls & \scriptsize PredCls & & \scriptsize SGCls & \scriptsize PredCls & & \scriptsize SGCls & \scriptsize PredCls\Tstrut\\
			\toprule
			Baseline (IMP++) & 9.3\std{0.1} & 28.1\std{0.1} & & 27.8\std{0.1} & 33.7\std{0.3} & & 48.7\std{0.1} & 77.5\std{0.1}\Bstrut\\
			Baseline, $\times 2$ updates & 8.2 & 25.8 & & 27.8 & 35.1 & & 46.6 & 73.5 \\
			\bottomrule
		\end{tabular}
		\vspace{-5pt}
	\end{table}

	\subsection{SGGen/SGDet Results\label{sec:sggen}}
	In addition to SGCls/PredCls results presented in the main text, we follow previous work~\cite{zellers2018neural,tang2020unbiased} and provide \textbf{SGGen} (SGDet) results that rely on using the bounding boxes predicted by the detector as opposed to using GT boxes (Table~\ref{tab:sggen_results}).
	We follow a standard refinement procedure~\cite{zellers2018neural,knyazev2020graph} and fine-tune SGCls models with $bs=6$ and $lr=(1e-4) \times bs$. We fine-tune for 2 epochs in all cases, which we set heuristically as our main goal in this task is not to outperform state-of-the-art, but to compare our model with the baseline. During this fine-tuning for simplicity we use only the baseline loss for all models, so that this step's main purpose is to adapt the SGCls model to predicted bounding boxes instead of ground truth. Despite these simplifications our GAN model still improves on the baseline. However, the results are worse than in previous works, perhaps because: (1) they rely on a stronger detector; (2) we do not update the detector on the generated features conditioned on rare compositions. Addressing these limitations can be the focus of future work. The models with reweighted predicates (\S~\ref{sec:reweight}) significantly improve the results on mean recall (as expected), and interestingly, on one of the zero-shot metrics.
	\looseness-1

	\begin{table}[t]
		\setlength{\tabcolsep}{2pt}
		\footnotesize
		\begin{center}
			\vspace{-5pt}
			\caption{SGGen results. The top-1 result in each column is \textbf{bolded}.
$^\dagger$The results in~\cite{tang2020unbiased} are obtained with a more advanced detector and, thus, are not directly comparable.}\label{tab:sggen_results}
			\vspace{1pt}
			\begin{tabular}{p{2.7cm}ccp{0.3cm}ccp{0.3cm}cc}
				\multirow{2}{*}{\textsc{\textbf{Model}}} & \multicolumn{2}{c}{\textsc{\textbf{\scriptsize Zero-shot}}} & & \multicolumn{2}{c}{\textsc{\textbf{\scriptsize Mean Recall}}} & & \multicolumn{2}{c}{\textsc{\textbf{\scriptsize All-Shot}}} \\
				 &
				\multicolumn{2}{c}{\tiny{SGGen \textbf{zsR}@100}} & & 
				\multicolumn{2}{c}{\tiny{SGGen \textbf{mR}@100}} & &
				\multicolumn{2}{c}{\tiny{SGGen \textbf{R}@100}}\\
				\cline{2-3}\cline{5-6}\cline{8-9}
				\multicolumn{1}{r}{\tiny{Graph Constraint}} & 
				\cmark & \xmark & &
				\cmark & \xmark & &
				\cmark & \xmark\Bstrut\Tstrut\\
				
				\toprule
				\textsc{Freq}~\cite{zellers2018neural,chen2019knowledge,knyazev2020graph} & 0.0 & 0.1 & & 5.6 & 8.9 & & 27.6 & 30.9\\
				IMP+~\cite{xu2017scene,zellers2018neural,knyazev2020graph}  & 0.9 & 0.9 & & 4.8 & 8.0 & & 24.5 & 27.4\\
				NM~\cite{zellers2018neural,chen2019knowledge,knyazev2020graph} & 0.3 & 0.8 & & 6.1 & 12.9 & & 30.3 & \textbf{35.8}\\
				KERN~\cite{chen2019knowledge,knyazev2020graph} & 0.0 & 0.0 & & 7.3 & \textbf{16.0} & & 29.8 & \textbf{35.8}\\
				VCTree$^\dagger$~\cite{tang2020unbiased} & 0.7 & $-$ & & 6.9 & $-$ & & 36.2 & $-$\\
				NM$^\dagger$~\cite{tang2020unbiased} & 0.2 & $-$ & & 6.8 & $-$ & & \textbf{36.9} & $-$\\
				NM+TDE$^\dagger$~\cite{tang2020unbiased} & \textbf{2.9} & $-$ & & 9.8 & $-$ & & 20.3 & $-$\\
				\hline
				Baseline (IMP++) & 0.8 & 0.9 & & 5.9 & 9.8 & & 25.1 & 28.1\Tstrut\\
				GAN & 1.0 & 1.2 & & 6.2 & 10.5 & & 25.4 & 28.6 \\
				GAN$\times w^2$ & 1.1 & \textbf{2.2} & & \textbf{9.9} & 15.9 & & 18.5 & 24.0 \\
				
				\bottomrule
			\end{tabular}
		\end{center}
		\vspace{-15pt}
	\end{table}

\subsection{Evaluation of Generated Visual Features\label{apdx:gan}}	

\begin{figure}[t]
\centering
\setlength{\tabcolsep}{1.5pt}
\begin{tabular}{cc}
     \footnotesize \textbf{\textsc{Real node features}} $V$ & \footnotesize \textbf{\textsc{Fake node features}} $\hat{V}$ \\
     \includegraphics[width=0.22\textwidth]{tsne_gan_nodes_test_zs_real.pdf} & \includegraphics[width=0.22\textwidth]{tsne_gan_nodes_test_zs_fake.pdf} \\
     \multicolumn{2}{c}{{\includegraphics[width=0.46\textwidth, trim={3cm 8.5cm 0.5cm 0.5cm}, clip]{tsne_gan_nodes_test_zs_legend.pdf}}} \\
     \bf\footnotesize \textsc{Real edge features} & \bf\footnotesize \textsc{Fake edge features} \\
     \includegraphics[width=0.22\textwidth]{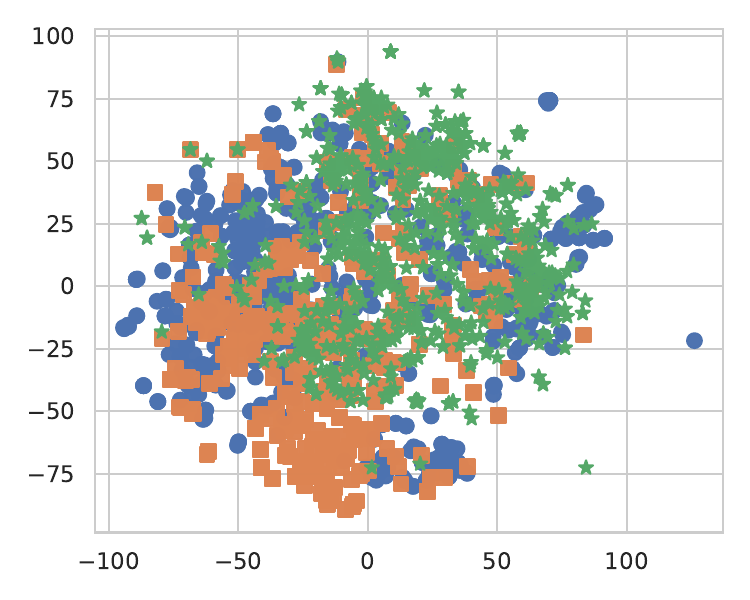} & \includegraphics[width=0.22\textwidth]{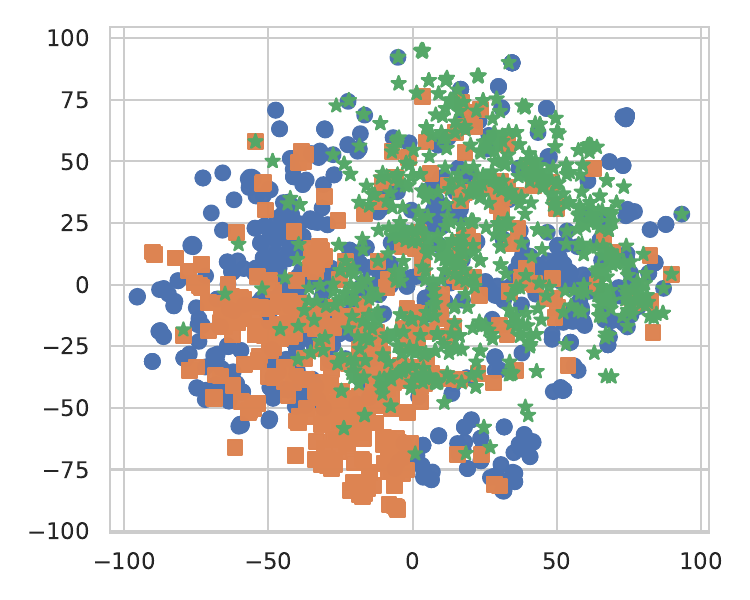} \\
     \multicolumn{2}{c}{{\includegraphics[width=0.35\textwidth, trim={9.5cm 8.5cm 0.5cm 0.5cm}, clip]{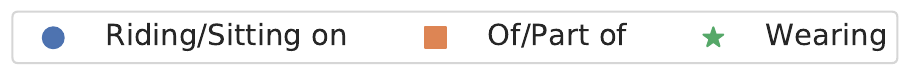}}} \\
     
     \multicolumn{2}{c}{\bf\footnotesize \textsc{Global features}}
     \\
    \multicolumn{2}{c}{\includegraphics[width=0.22\textwidth]{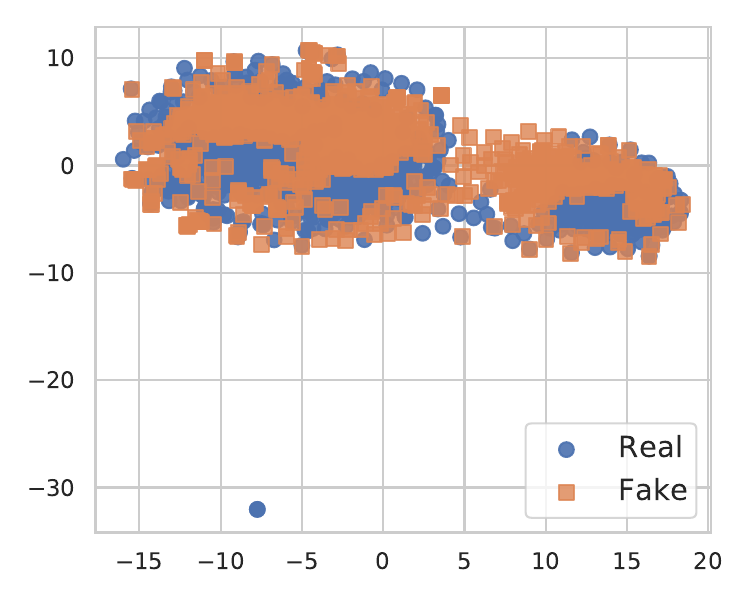}} \\    
\end{tabular}
  \caption{\small Visualizations of real \textit{vs} generated node, edge and global features obtained using t-SNE.}
\label{fig:tsne_more}
\end{figure}

In addition to the quality estimation of node features in the main text, we evaluate edge and global features qualitatively (Fig.~\ref{fig:tsne_more}) and quantitatively (Table~\ref{tab:gen_more}).
We also visualize features by averaging over the channel dimension (Fig.~\ref{fig:fmaps},~\ref{fig:node_edge}). These visualizations show that the generated samples are diverse and respond to the changes in the conditioning. However, the global features generated by our GAN are noticeably more smooth, which might be an effect of the generator's architectural inductive bias or visualization strategy.\looseness=-1

\begin{table}[b]
    \vspace{-10pt}
	\caption{Evaluation of generated (fake) feature using the metrics of ``similarity'' between two distributions $X$ and $Y$~\cite{kynkaanniemi2019improved,naeem2020reliable}. The same held-out set of real test features ($Y \sim V$) is used as the reference distribution in all cases. The percentage in the superscripts denotes a relative drop of the average metric when switching from test to test-zs conditioning. For all metrics, higher is better.\looseness-1}
	\label{tab:gen_more}
	\vspace{1pt}
	\centering
	\scriptsize
	\setlength{\tabcolsep}{1.5pt}
	\begin{tabular}{p{0.2cm}l|cc|cc|p{1.2cm}}
	\toprule
	    & \multirow{2}{*}{\tiny\bf\textsc{Distribution $X$}} & \multicolumn{2}{c|}{\bf Fidelity (realism)} & \multicolumn{2}{c|}{\bf Diversity} & \multicolumn{1}{c}{\multirow{2}{*}{\bf \textsc{Avg}}} \\
		& & \bf \textsc{Precision} & \bf \textsc{Density} 
		& \bf \textsc{Recall} & \bf \textsc{Coverage} & \Bstrut\\
		\hline
		\multirow{4}{*}{\rotatebox[origin=c]{90}{\textbf{\textsc{Nodes}}}}
		& Real test & 0.74 & 1.02 & 0.75 & 0.97 & 0.87 \Tstrut\\
		& Real test-zs & 0.66 & 0.99 & 0.70 & 0.94 & 0.82$^{-6\%}$ \\
		& GAN: Fake test & 0.55 & 0.77 & 0.42 & 0.82 & 0.64 \\
		& GAN: Fake test-zs & 0.47 & 0.60 & 0.41 & 0.75 & 0.56$^{-13\%}$\Bstrut\\
        \hline
        \multirow{4}{*}{\rotatebox[origin=c]{90}{\textbf{\textsc{Edges}}}}
        & Real test & 0.73 & 0.97 & 0.72 & 0.97 & 0.85\Tstrut\\
        & Real test-zs & 0.53& 0.99 & 0.59  & 0.87 & 0.75$^{-12\%}$\\
        & GAN: Fake test & 0.54 & 0.59 & 0.50 & 0.75 & 0.60 \\
        & GAN: Fake test-zs & 0.38 & 0.36 & 0.50 & 0.58 & 0.46$^{-23\%}$\Bstrut\\
        \hline
        \multirow{4}{*}{\rotatebox[origin=c]{90}{\textbf{\textsc{Global}}}} & 
        Real test & 0.30 & 0.96 & 0.31 & 0.99 & 0.64\Tstrut\\
        & Real test-zs & 0.20 & 0.91 & 0.35 & 0.73 & 0.55$^{-14\%}$\\
        & GAN: Fake test & 0.14 & 0.43 & 0.22 & 0.83 & 0.41 \\
        & GAN: Fake test-zs & 0.11 & 0.23 & 0.29 & 0.68 & 0.33$^{-20\%}$ \\
	    \bottomrule
    \end{tabular}
    \vspace{0pt}
\end{table}

\begin{figure}
\centering
\newcommand{\width}{0.18\textwidth}
\newcommand{\height}{2cm}
\setlength{\tabcolsep}{2pt}
\scriptsize
\begin{tabular}{ccc}
     & Input image & Real feature map $H$ \\
     & \includegraphics[width=\width]{2350517_sup.png} & {\includegraphics[width=\width,trim={2.8cm 3.2cm 2.8cm 0.5cm},clip]{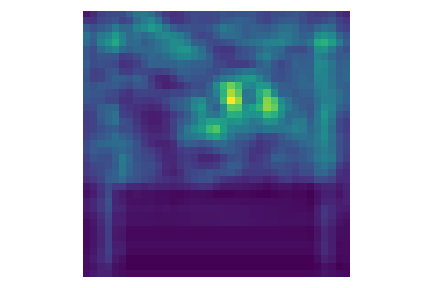}}  \\
     \hline \\
     & Generated $\hat{H}$ (sample \# 1) & Generated $\hat{H}$ (sample \# 2) \\
     \rotatebox[origin=c]{90}{$\graph$} & {\includegraphics[width=\width,trim={2.8cm 3.2cm 2.8cm 0.5cm},clip,align=c]{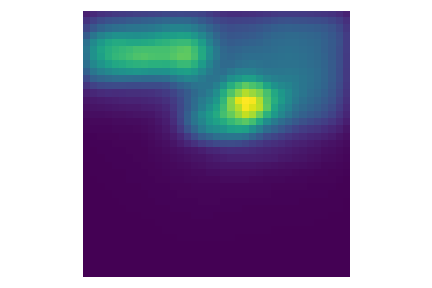}} & {\includegraphics[width=\width,trim={2.8cm 3.2cm 2.8cm 0.5cm},clip,align=c]{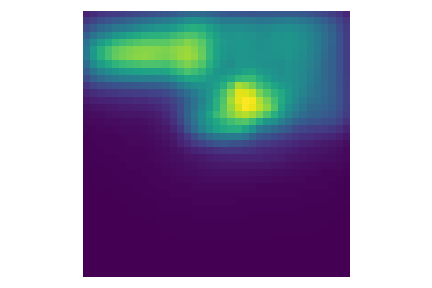}}  \\
     \rotatebox[origin=c]{90}{$\pgraph$: person $\rightarrow$ dog} & \includegraphics[width=\width,trim={2.8cm 3.2cm 2.8cm 0.5cm},clip,align=c]{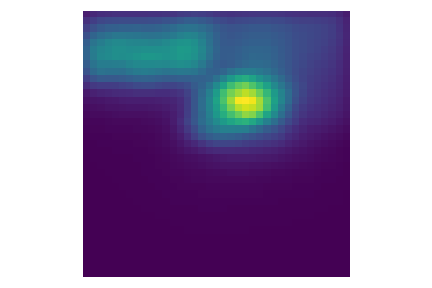} & {\includegraphics[width=\width,trim={2.8cm 3.2cm 2.8cm 0.5cm},clip,align=c]{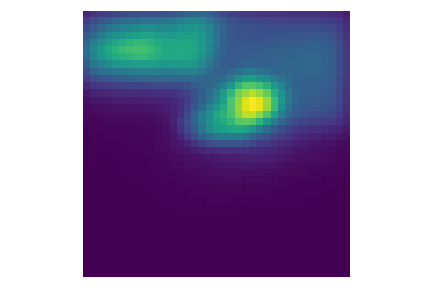}}  \\
\end{tabular}
\vspace{3pt}
\caption{Real $H$ and Fake (generated) $\hat{H}$ global feature maps (averaged over the channel dimension) with a perturbation for the triplet ``person on surfboard'' in $\graph$.}
\label{fig:fmaps}
\vspace{-10pt}
\end{figure}

\begin{figure}
    \vspace{-5pt}
    \centering
    \newcommand{\width}{0.05\textwidth}
    \newcommand{\height}{2cm}
    \setlength{\tabcolsep}{2pt}
    \footnotesize
    \begin{tabular}{ccccp{0.3cm}ccc}
        & \multicolumn{3}{c}{Features of objects \textbf{Person}} & & \multicolumn{3}{c}{Features of predicates \textbf{On}}\\
        {\rotatebox[origin=c]{90}{Real}} & \includegraphics[width=\width,trim={2.8cm 0.1cm 2.8cm 0.1cm},clip,align=c]{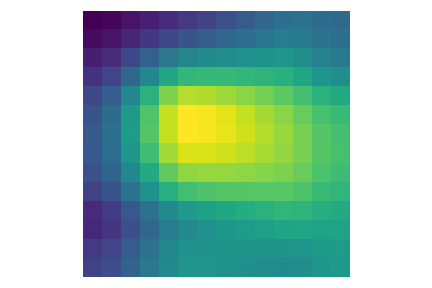} &
        \includegraphics[width=\width,trim={2.8cm 0.1cm 2.8cm 0.1cm},clip,align=c]{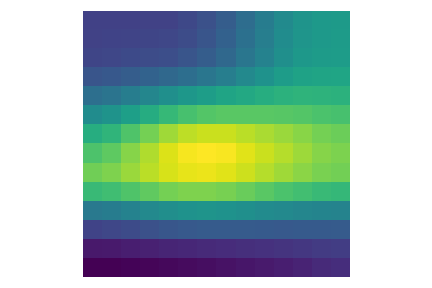} &
        \includegraphics[width=\width,trim={2.8cm 0.1cm 2.8cm 0.1cm},clip,align=c]{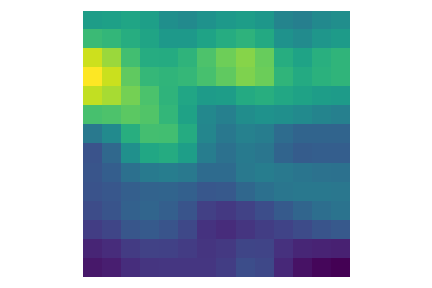} & &
        \includegraphics[width=\width,trim={2.8cm 0.1cm 2.8cm 0.1cm},clip,align=c]{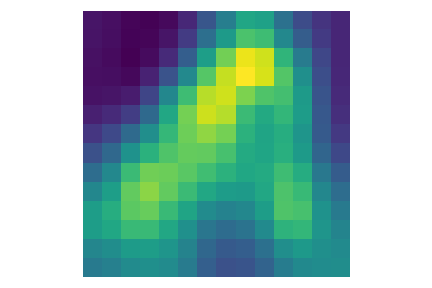} &
        \includegraphics[width=\width,trim={2.8cm 0.1cm 2.8cm 0.1cm},clip,align=c]{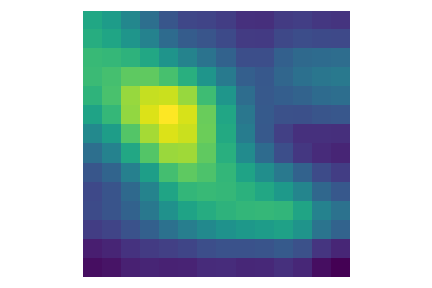} &
        \includegraphics[width=\width,trim={2.8cm 0.1cm 2.8cm 0.1cm},clip,align=c]{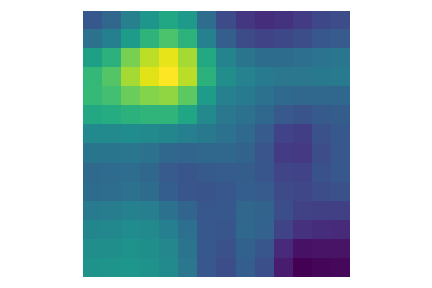} \\
        {\rotatebox[origin=c]{90}{Fake}} & \includegraphics[width=\width,trim={2.8cm 0.1cm 2.8cm 0.1cm},clip,align=c]{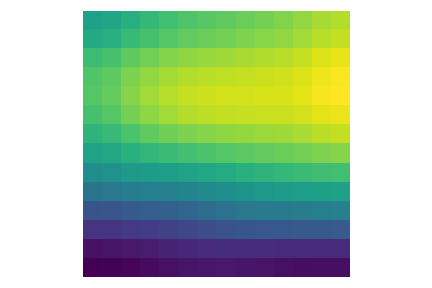} &
        \includegraphics[width=\width,trim={2.8cm 0.1cm 2.8cm 0.1cm},clip,align=c]{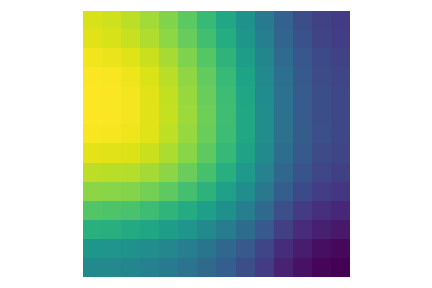} &
        \includegraphics[width=\width,trim={2.8cm 0.1cm 2.8cm 0.1cm},clip,align=c]{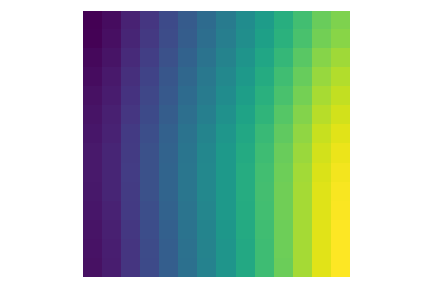} & &
        \includegraphics[width=\width,trim={2.8cm 0.1cm 2.8cm 0.1cm},clip,align=c]{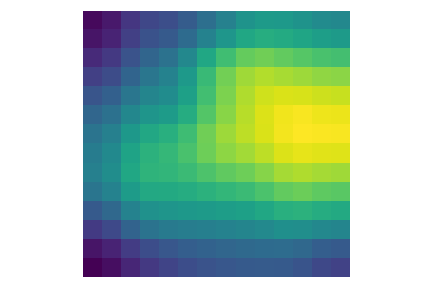} &
        \includegraphics[width=\width,trim={2.8cm 0.1cm 2.8cm 0.1cm},clip,align=c]{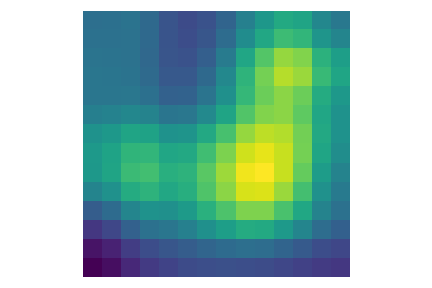} &
        \includegraphics[width=\width,trim={2.8cm 0.1cm 2.8cm 0.1cm},clip,align=c]{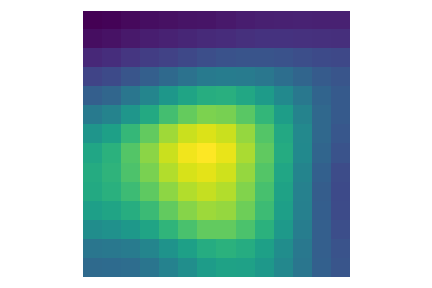} \\
    \end{tabular}
    \caption{Real and Fake node and edge features (averaged over the channel dimension) for objects and predicates.}
    \label{fig:node_edge}
    \vspace{-10pt}
\end{figure}

\subsection{Limitations\label{sec:nolimit}}
\vspace{-3pt}
Our method is limited in three main aspects. \textbf{First}, we rely on a pretrained object detector to extract visual features. Without generating augmentations all the way to the images --- in order to update the detector on rare compositions --- it is hard to obtain significantly stronger performance. While augmentations in the feature space can be effective~\cite{devries2017dataset, verma2019manifold}, their adoption for large-scale out-of-distribution generalization is underexplored. 
\textbf{Second}, by making a simplification and keeping GT bounding boxes for perturbed scene graphs, we limit (1) the amount of perturbations we can make (if we permit many nodes to be perturbed, then it is hard to expect the same layout), and (2) the diversity of spatial compositions, which might be an important aspect of compositional generalization. 
We attempted to verify that using \oracle~perturbations, which are created by directly using ZS triplets from the test set.
Using \oracle~with GT bounding boxes (our default setting) surprisingly does not result in large improvements. However, when we replace GT boxes with the ones taken from the corresponding samples of the test set, the results improve significantly.
This demonstrates that: (1) our GAN model may benefit from reliable bounding box prediction (e.g.~\cite{hong2018inferring}); (2) \structn~perturbations are already effective (close to \oracle) and improving the results further by relying solely on perturbations is challenging.
\textbf{Third}, the quality of generated features, especially, for novel and rare compositions is currently limited, which is also carefully analyzed in~\cite{casanova2020generating}. Addressing this challenge can further improve results both of \oracle~and non-\oracle~models.
\looseness-1	
	
\section{Additional visualizations\label{sec:vis}}
\begin{figure*}[thpb]
    \centering
    \vspace{-10pt}
    \small
    \newcommand{\width}{0.28\textwidth}
    \newcommand{\height}{3cm}
    \setlength{\tabcolsep}{30pt}
    \begin{tabular}{cc}
         \includegraphics[width=\width,height=\height]{2350517_sup.png} &  \includegraphics[width=0.25\textwidth,trim={1.5cm 0.1cm 7cm 0.1cm},clip]{2350517_gt_graph_sup.png} \\
     \end{tabular}
     \setlength{\tabcolsep}{2pt}
     \begin{tabular}{|c|c|cc|}
         \toprule
         \textbf{\textsc{Rand}} & \textbf{\textsc{Neigh}} & \textbf{\structn} & \\
         \hline
         \includegraphics[width=\width,align=c,trim={1.5cm 0.1cm 7cm 0.1cm},clip]{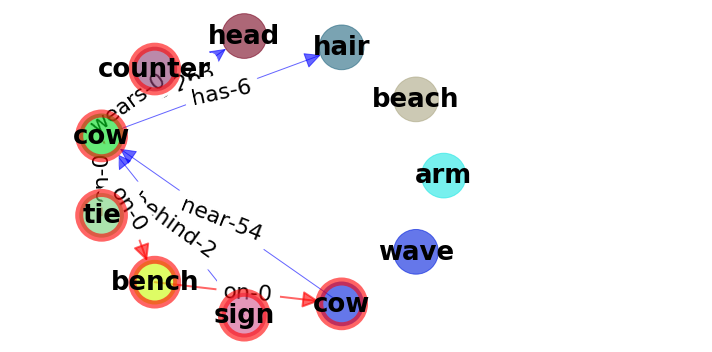} &
         \includegraphics[width=\width,align=c,trim={1.5cm 0.1cm 7cm 0.1cm},clip]{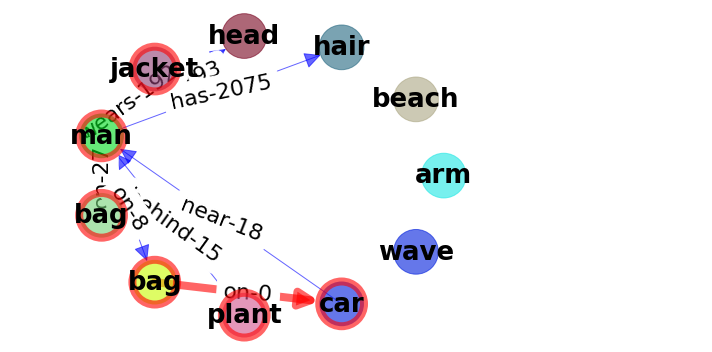} & 
         \includegraphics[width=\width,align=c,trim={1.5cm 0.1cm 7cm 0.1cm},clip]{2350517_structn_L0_5_topk5_a2_graph_13.png} & {\rotatebox[origin=c]{90}{$\alpha=2$}} \\
         
         \includegraphics[width=\width,align=c,trim={1.5cm 0.1cm 7cm 0.1cm},clip]{2350517_rand_L0_5_topk10_a2_graph_1.png} &
         \includegraphics[width=\width,align=c,trim={1.5cm 0.1cm 7cm 0.1cm},clip]{2350517_neigh_L0_5_topk10_a2_graph_1.png} & \includegraphics[width=\width,align=c,trim={1.5cm 0.1cm 7cm 0.1cm},clip]{2350517_structn_L0_5_topk5_a5_graph_3.png} & {\rotatebox[origin=c]{90}{$\alpha=5$}} \\
         
         \includegraphics[width=\width,align=c,trim={1.5cm 0.1cm 7cm 0.1cm},clip]{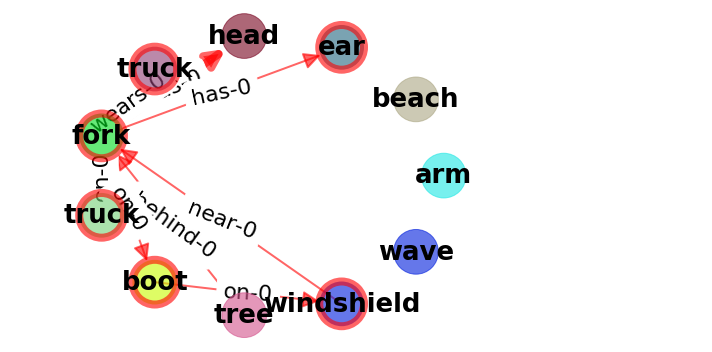} &
         \includegraphics[width=\width,align=c,trim={1.5cm 0.1cm 7cm 0.1cm},clip]{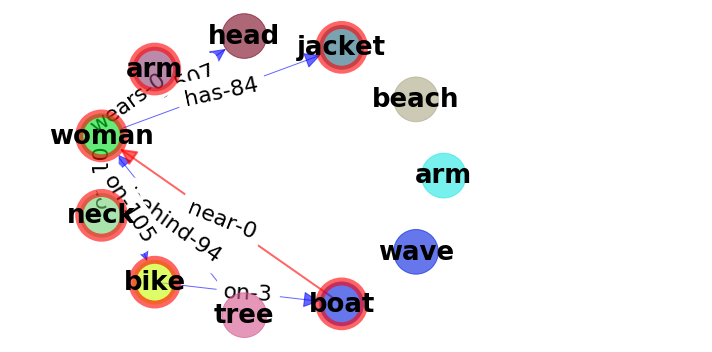} & \includegraphics[width=\width,align=c,trim={1.5cm 0.1cm 7cm 0.1cm},clip]{2350517_structn_L0_5_topk5_a10_graph_7.png} & {\rotatebox[origin=c]{90}{$\alpha=10$}} \\
         
         \includegraphics[width=\width,align=c,trim={1.5cm 0.1cm 7cm 0.1cm},clip]{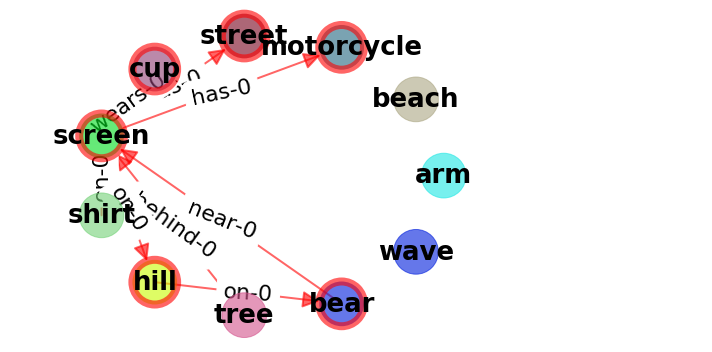} &
         \includegraphics[width=\width,align=c,trim={1.5cm 0.1cm 7cm 0.1cm},clip]{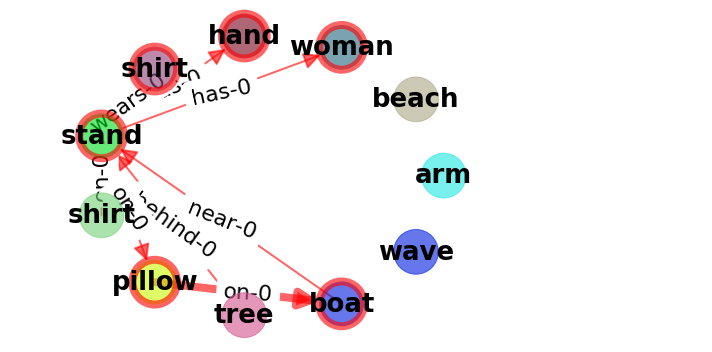} & \includegraphics[width=\width,align=c,trim={1.5cm 0.1cm 7cm 0.1cm},clip]{2350517_structn_L0_5_topk5_a20_graph_2.png} & {\rotatebox[origin=c]{90}{$\alpha=20$}}\\
         
         \includegraphics[width=\width,align=c,trim={1.5cm 0.1cm 7cm 0.1cm},clip]{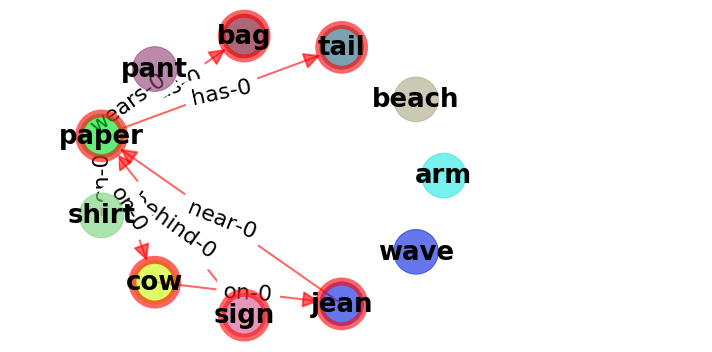} &
         \includegraphics[width=\width,align=c,trim={1.5cm 0.1cm 7cm 0.1cm},clip]{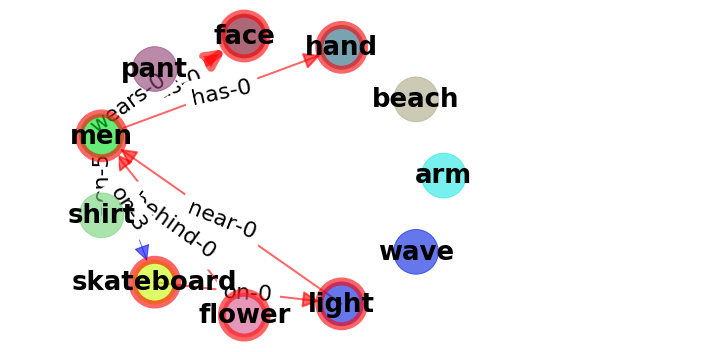} & \includegraphics[width=\width,align=c,trim={1.5cm 0.1cm 7cm 0.1cm},clip]{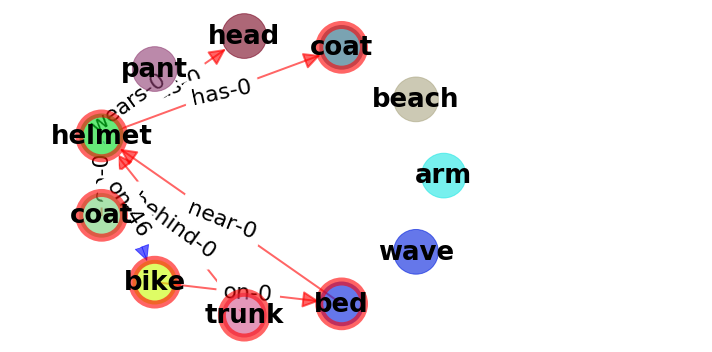} & {\rotatebox[origin=c]{90}{$\alpha=50$}}\\
         \bottomrule
    \end{tabular}
    \vspace{5pt}
    \caption{Visualizations of perturbations for image \texttt{2350517} from Visual Genome.}
    \label{fig:examples1}
\end{figure*}

\begin{figure*}[thpb]
    \centering
    \vspace{-10pt}
    \small
    \newcommand{\width}{0.28\textwidth}
    \newcommand{\height}{3cm}
    \setlength{\tabcolsep}{30pt}
    \begin{tabular}{cc}
         \includegraphics[width=\width,height=\height]{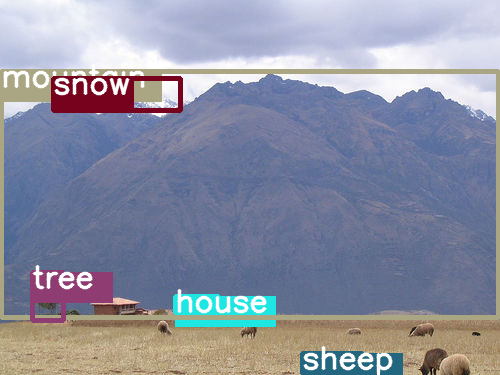} &  \includegraphics[width=0.25\textwidth,trim={1.5cm 0.1cm 7cm 0.1cm},clip]{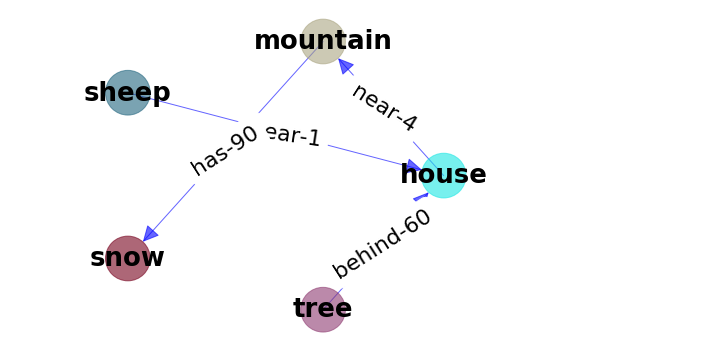} \\
     \end{tabular}
     \setlength{\tabcolsep}{2pt}
     \begin{tabular}{|c|c|cc|}
         \toprule
         \textbf{\textsc{Rand}} & \textbf{\textsc{Neigh}} & \textbf{\structn} & \\
         \hline
         \includegraphics[width=\width,align=c,trim={1.5cm 0.1cm 7cm 0.1cm},clip]{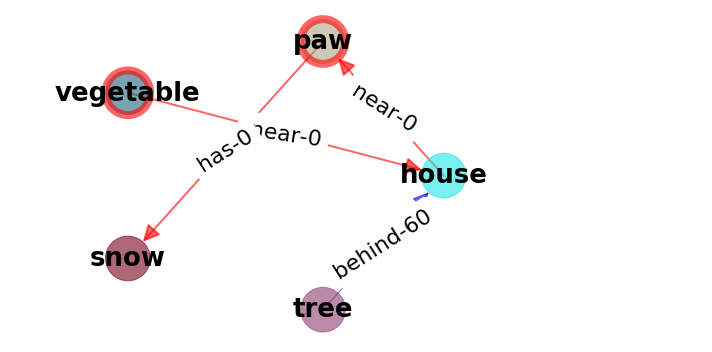} &
         \includegraphics[width=\width,align=c,trim={1.5cm 0.1cm 7cm 0.1cm},clip]{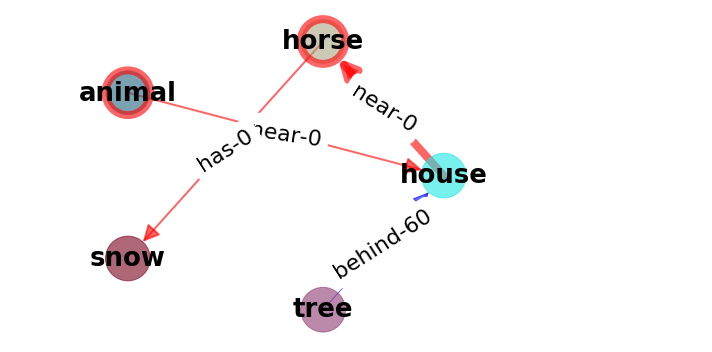} & 
         \includegraphics[width=\width,align=c,trim={1.5cm 0.1cm 7cm 0.1cm},clip]{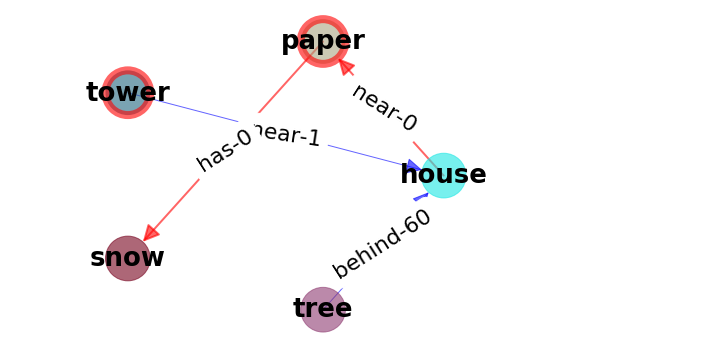} & {\rotatebox[origin=c]{90}{$\alpha=2$}} \\
         
         \includegraphics[width=\width,align=c,trim={1.5cm 0.1cm 7cm 0.1cm},clip]{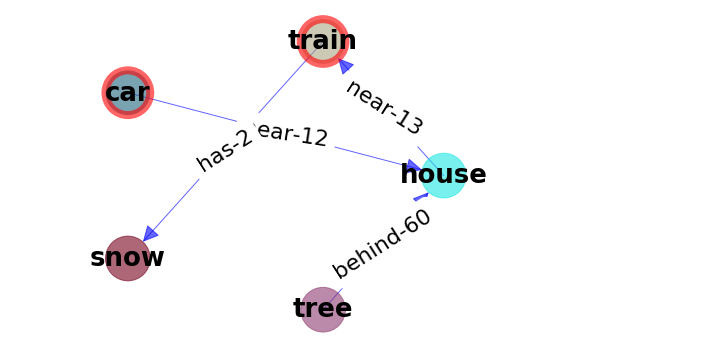} &
         \includegraphics[width=\width,align=c,trim={1.5cm 0.1cm 7cm 0.1cm},clip]{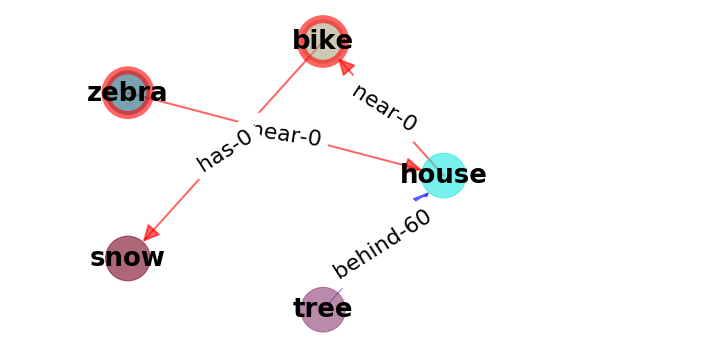} & \includegraphics[width=\width,align=c,trim={1.5cm 0.1cm 7cm 0.1cm},clip]{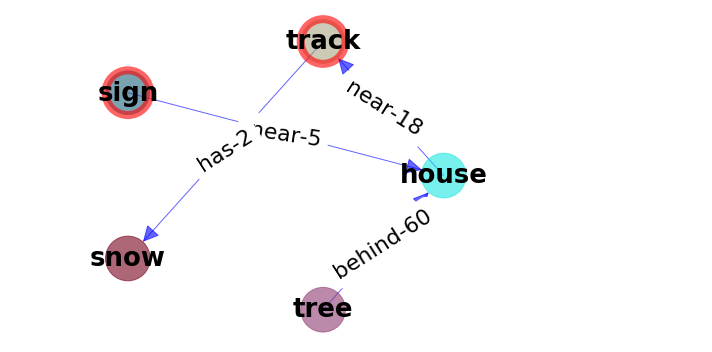} & {\rotatebox[origin=c]{90}{$\alpha=5$}} \\

         \includegraphics[width=\width,align=c,trim={1.5cm 0.1cm 7cm 0.1cm},clip]{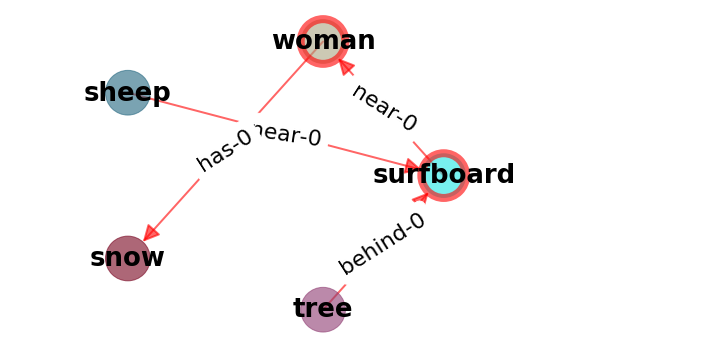} &
         \includegraphics[width=\width,align=c,trim={1.5cm 0.1cm 7cm 0.1cm},clip]{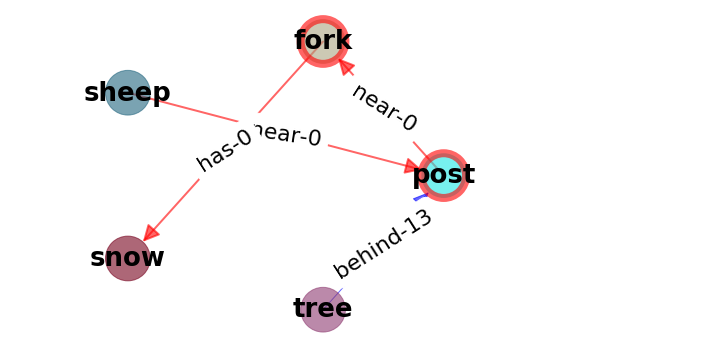} & \includegraphics[width=\width,align=c,trim={1.5cm 0.1cm 7cm 0.1cm},clip]{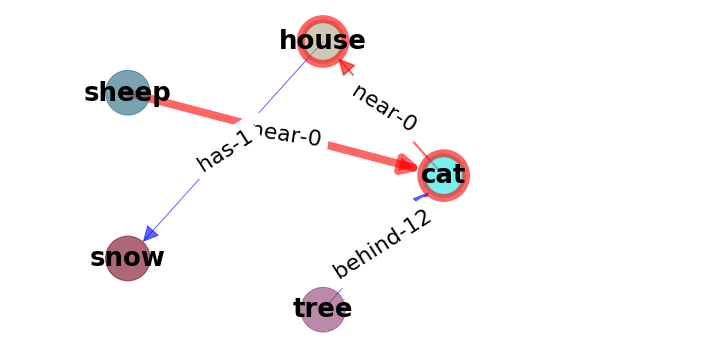} & {\rotatebox[origin=c]{90}{$\alpha=10$}} \\

         \includegraphics[width=\width,align=c,trim={1.5cm 0.1cm 7cm 0.1cm},clip]{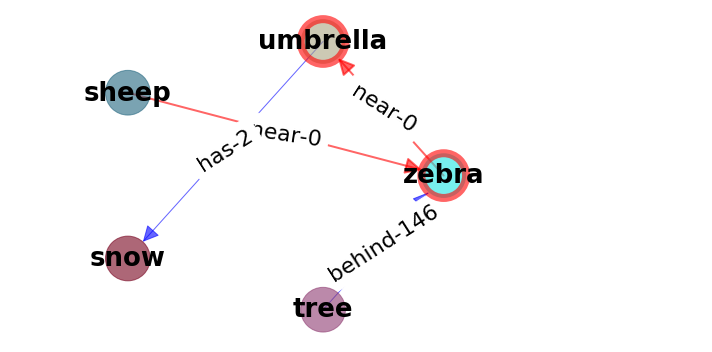} &
         \includegraphics[width=\width,align=c,trim={1.5cm 0.1cm 7cm 0.1cm},clip]{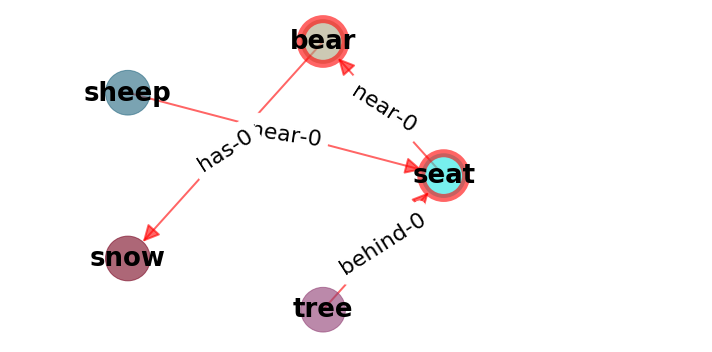} & \includegraphics[width=\width,align=c,trim={1.5cm 0.1cm 7cm 0.1cm},clip]{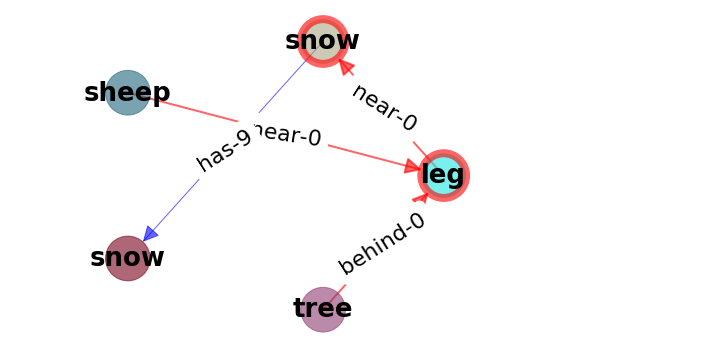} & {\rotatebox[origin=c]{90}{$\alpha=20$}}\\
         
         \includegraphics[width=\width,align=c,trim={1.5cm 0.1cm 7cm 0.1cm},clip]{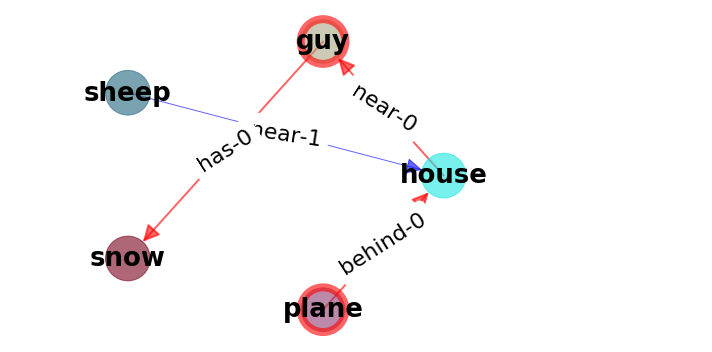} &
         \includegraphics[width=\width,align=c,trim={1.5cm 0.1cm 7cm 0.1cm},clip]{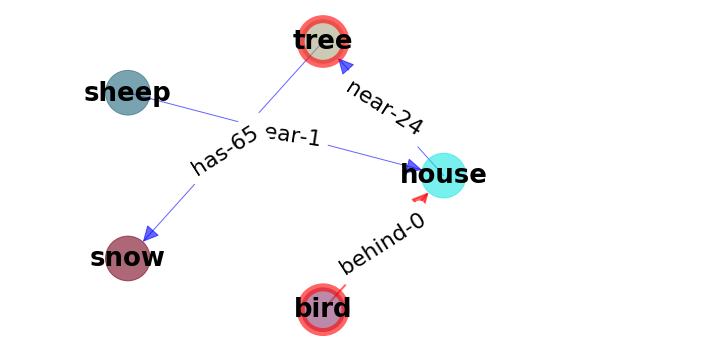} & \includegraphics[width=\width,align=c,trim={1.5cm 0.1cm 7cm 0.1cm},clip]{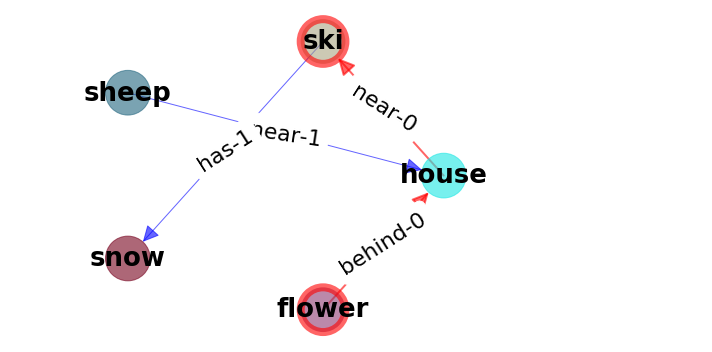} & {\rotatebox[origin=c]{90}{$\alpha=50$}} \\
         \bottomrule
    \end{tabular}
    \vspace{5pt}
    \caption{Visualizations of perturbations for image \texttt{2343590} from Visual Genome.}
    \label{fig:examples2}
\end{figure*}

\begin{figure*}[thpb]
    \centering
    \vspace{-10pt}
    \small
    \newcommand{\width}{0.28\textwidth}
    \newcommand{\height}{3cm}
    \setlength{\tabcolsep}{30pt}
    \begin{tabular}{cc}
         \includegraphics[width=0.22\textwidth,height=\height]{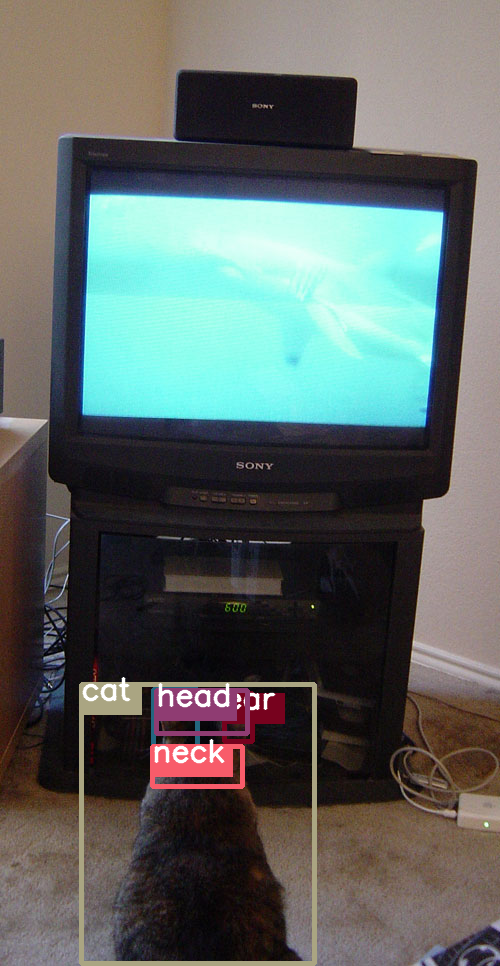} &  \includegraphics[width=0.25\textwidth,trim={1.5cm 0.1cm 7cm 0.1cm},clip]{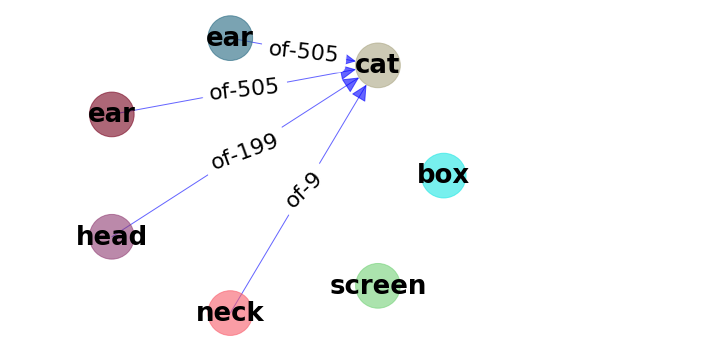} \\
     \end{tabular}
     \setlength{\tabcolsep}{2pt}
     \begin{tabular}{|c|c|cc|}
         \toprule
         \textbf{\textsc{Rand}} & \textbf{\textsc{Neigh}} & \textbf{\structn} & \\
         \toprule
         \includegraphics[width=\width,align=c,trim={1.5cm 0.1cm 7cm 0.1cm},clip]{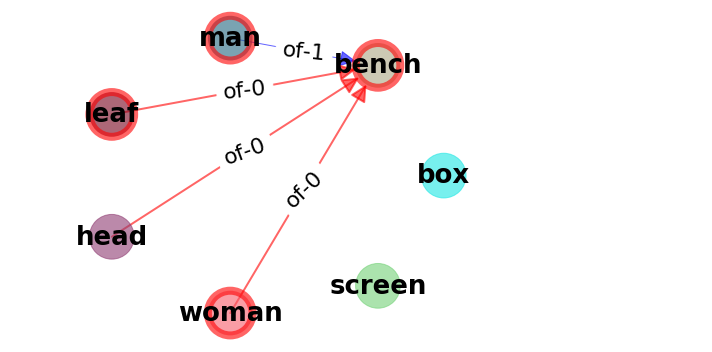} &
         \includegraphics[width=\width,align=c,trim={1.5cm 0.1cm 7cm 0.1cm},clip]{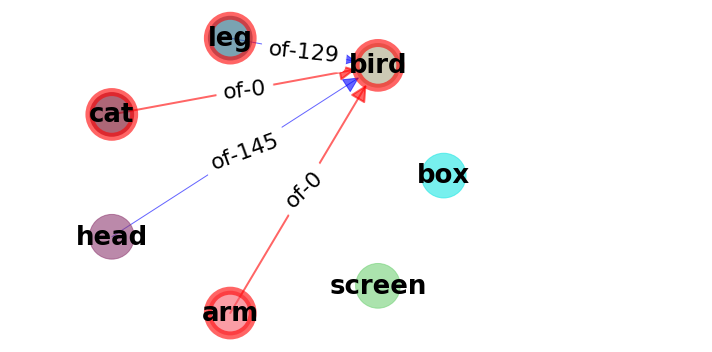} & 
         \includegraphics[width=\width,align=c,trim={1.5cm 0.1cm 7cm 0.1cm},clip]{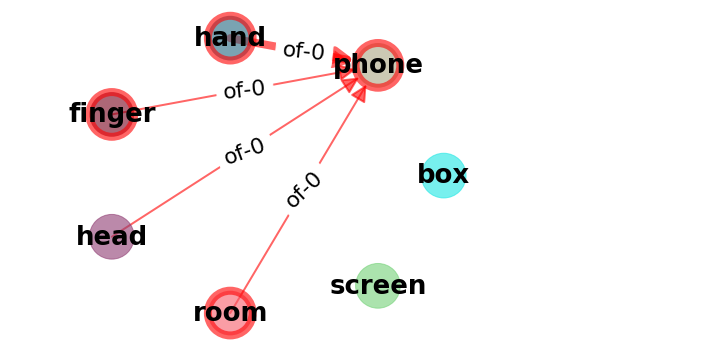} & {\rotatebox[origin=c]{90}{$\alpha=2$}} \\
         
         \includegraphics[width=\width,align=c,trim={1.5cm 0.1cm 7cm 0.1cm},clip]{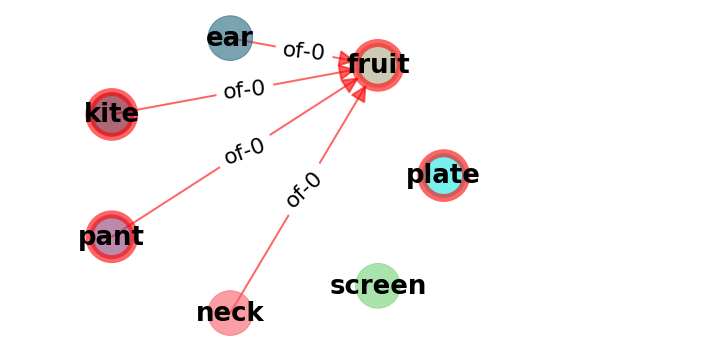} &
         \includegraphics[width=\width,align=c,trim={1.5cm 0.1cm 7cm 0.1cm},clip]{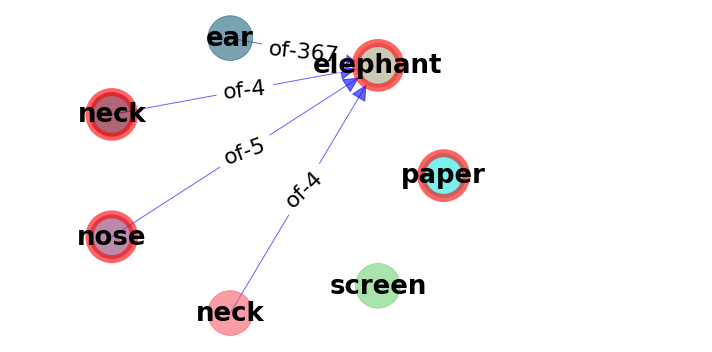} & \includegraphics[width=\width,align=c,trim={1.5cm 0.1cm 7cm 0.1cm},clip]{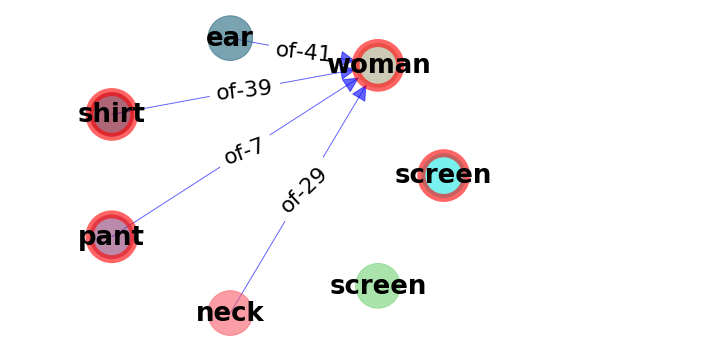} & {\rotatebox[origin=c]{90}{$\alpha=5$}} \\

         \includegraphics[width=\width,align=c,trim={1.5cm 0.1cm 7cm 0.1cm},clip]{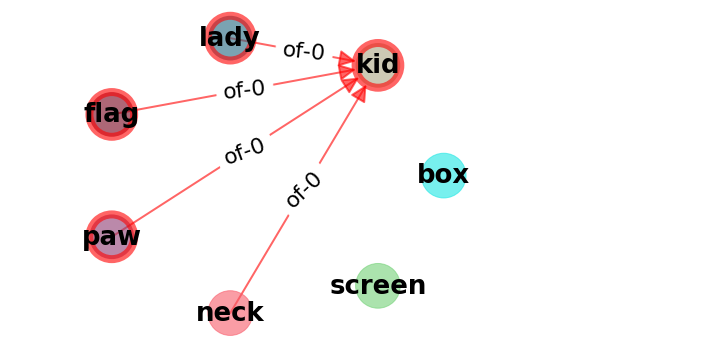} &
         \includegraphics[width=\width,align=c,trim={1.5cm 0.1cm 7cm 0.1cm},clip]{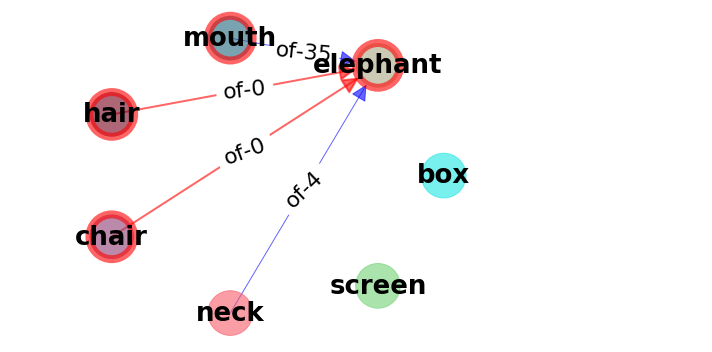} & \includegraphics[width=\width,align=c,trim={1.5cm 0.1cm 7cm 0.1cm},clip]{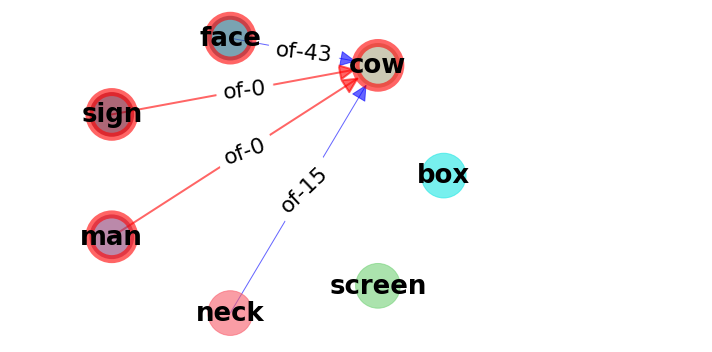} & {\rotatebox[origin=c]{90}{$\alpha=10$}} \\

         \includegraphics[width=\width,align=c,trim={1.5cm 0.1cm 7cm 0.1cm},clip]{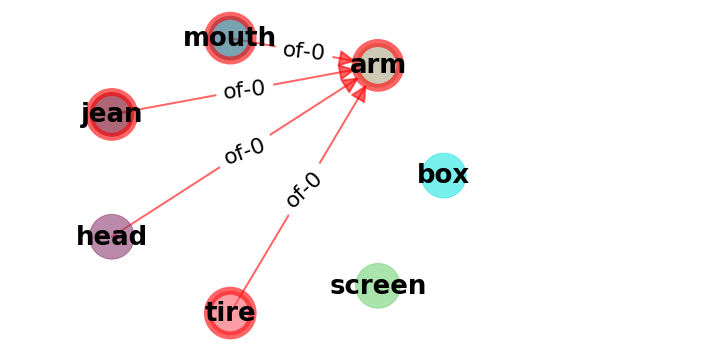} &
         \includegraphics[width=\width,align=c,trim={1.5cm 0.1cm 7cm 0.1cm},clip]{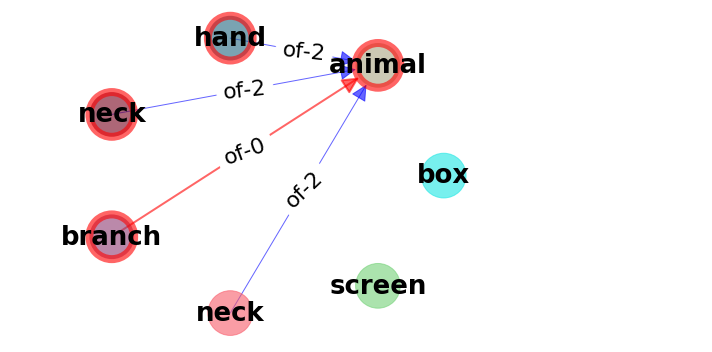} & \includegraphics[width=\width,align=c,trim={1.5cm 0.1cm 7cm 0.1cm},clip]{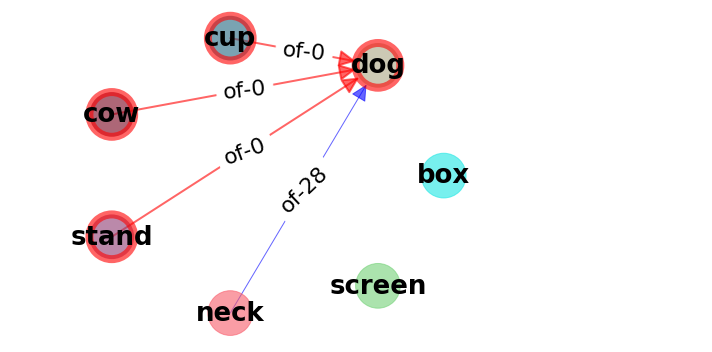} & {\rotatebox[origin=c]{90}{$\alpha=20$}}\\
         
         \includegraphics[width=\width,align=c,trim={1.5cm 0.1cm 7cm 0.1cm},clip]{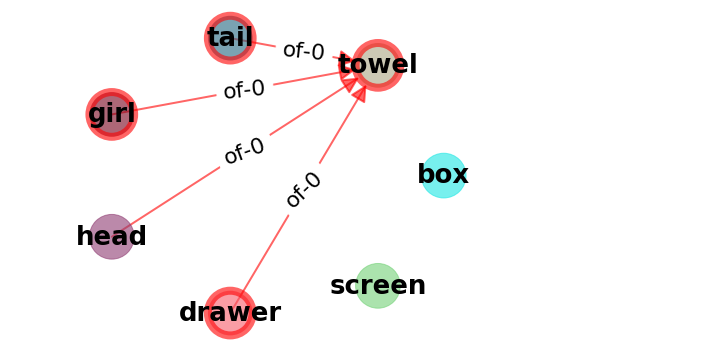} &
         \includegraphics[width=\width,align=c,trim={1.5cm 0.1cm 7cm 0.1cm},clip]{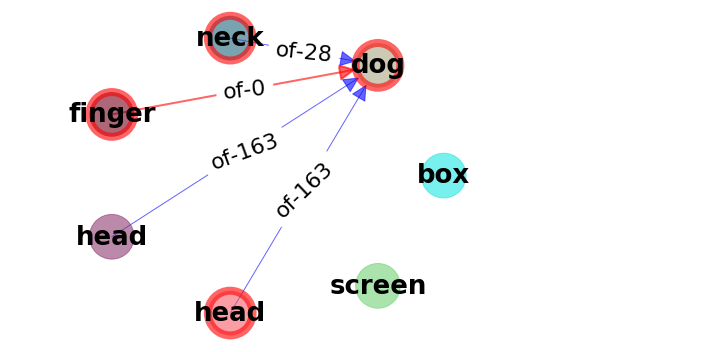} & \includegraphics[width=\width,align=c,trim={1.5cm 0.1cm 7cm 0.1cm},clip]{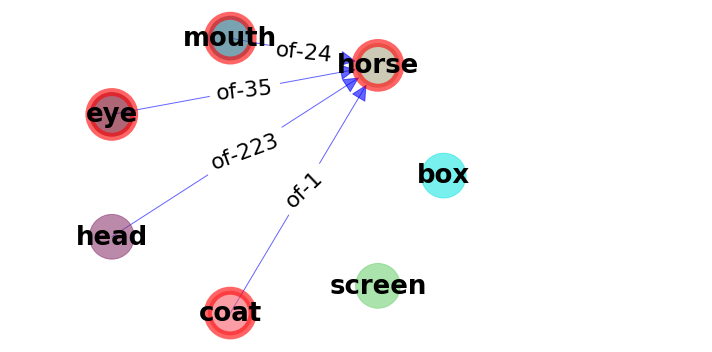} & {\rotatebox[origin=c]{90}{$\alpha=50$}} \\
         \bottomrule
    \end{tabular}
    \vspace{5pt}
    \caption{Visualizations of perturbations for image \texttt{1159620} from Visual Genome.}
    \label{fig:examples4}
\end{figure*}

\vspace{-3pt}
Figures~\ref{fig:examples1}-\ref{fig:examples4} show additional examples of different perturbations applied to scene graphs from Visual Genome (on top of each figure we show graph $\graph$ along with the corresponding image).
Perturbed nodes are highlighted in red. Red edges denote triplets missing both in the training and test sets. Thick red edges denote triplets only present in the test set, i.e. zero-shots. Blue edges denote triplets present in the training set, where the number indicates the total number of such triplets in the training set. These visualization show that in case of \textsc{Rand}, most of the created triplets are implausible as a result of random perturbations. \textsc{Neigh} leads to very likely compositions, but less often provides rare plausible compositions. In contrast, \structn~can create plausible compositions that are rare or more frequent depending on $\alpha$.

\end{document}